\documentclass[pdflatex,sn-mathphys-num,iicol]{sn-jnl}% Math and Physical Sciences Numbered Reference Style
\usepackage[utf8]{inputenc}

\usepackage{graphicx}%
\usepackage{multirow}%
\usepackage{amsmath,amssymb,amsfonts}%
\usepackage{amsthm}%
\usepackage{mathrsfs}%
\usepackage[title]{appendix}%
\usepackage{xcolor}%
\usepackage{textcomp}%
\usepackage{manyfoot}%
\usepackage{booktabs}%
\usepackage{algorithm}%
\usepackage{algorithmicx}%
\usepackage{algpseudocode}%
\usepackage{listings}%
\usepackage{tcolorbox}
\usepackage{wrapfig}
\usepackage{multirow}
\usepackage{booktabs} 
\usepackage{colortbl}
\newcommand{\xhdr}[1]{\textbf{#1}\:}

\usepackage{makecell}
\theoremstyle{thmstyleone}%
\theoremstyle{thmstyletwo}%

\theoremstyle{thmstylethree}%

\begin{document}

\title[Article Title]{Structural Pruning of Large Vision Language Models: A Comprehensive Study on Pruning Dynamics, Recovery, and Data Efficiency}


\author*[1,2,3]{\fnm{Yiran} \sur{Huang}}\email{yiran.huang@tum.de}
\author[3,4]{\fnm{Lukas} \sur{Thede}}\email{lukas.thede@uni-tuebingen.de}

\author[5]{\fnm{Massimiliano} \sur{Mancini}}\email{massimiliano.mancini@unitn.it}

\author[6]{\fnm{Wenjia} \sur{Xu}}\email{xuwenjia@bupt.edu.cn}
\author[1,2,3]{\fnm{Zeynep} \sur{Akata}}\email{zeynep.akata@tum.de}

\affil*[1]{\orgdiv{Technical University of Munich, Germany}
% \orgname{Organization}, \orgaddress{\street{Street}, \city{City}, \postcode{100190}, \state{State}, \country{Country}}
}
\affil[2]{\orgdiv{Munich Center for Machine Learning, Germany}}

\affil[3]{\orgdiv{Helmholtz Munich, Germany}}

\affil[4]{\orgdiv{University of Tübingen, Tübingen AI Center, Germany}}

\affil[5]{\orgdiv{University of Trento, Italy}}

\affil[6]{\orgdiv{Beijing University of Posts and Telecommunications, China}}


\abstract{While Large Vision Language Models (LVLMs) demonstrate impressive capabilities, their substantial computational and memory requirements pose deployment challenges on resource-constrained edge devices.
Current parameter reduction techniques primarily involve training LVLMs from small language models, but these methods offer limited flexibility and remain computationally intensive. 
We study a complementary route: compressing existing LVLMs by applying structured pruning to the language model backbone, followed by lightweight recovery training. 
Specifically, we investigate two structural pruning paradigms: layerwise and widthwise pruning, and pair them with supervised finetuning and knowledge distillation on logits and hidden states.
Additionally, we assess the feasibility of conducting recovery training with only a small fraction of the available data. 
Our results show that widthwise pruning generally maintains better performance in low-resource scenarios, where computational resources are limited or there is insufficient finetuning data. As for the recovery training, finetuning only the multimodal projector is sufficient at small compression levels. Furthermore, a combination of supervised finetuning and hidden-state distillation yields optimal recovery across various pruning levels. Notably, effective recovery can be achieved using just 5\% of the original data, while retaining over 95\% of the original performance. 
Through empirical study on three representative LVLM families ranging from 3B to 7B parameters, this study offers actionable insights for practitioners to compress LVLMs without extensive computation resources or sufficient data.
}




\maketitle

\section{Introduction}\label{sec1}
\begin{figure*}[h]
\definecolor{prune}{RGB}{134, 163, 210}
\definecolor{recovery}{RGB}{112, 175, 131}
\definecolor{scenario}{RGB}{118, 42, 131}
\definecolor{performance}{RGB}{255, 220, 100}
\centering
\includegraphics[width=\linewidth]{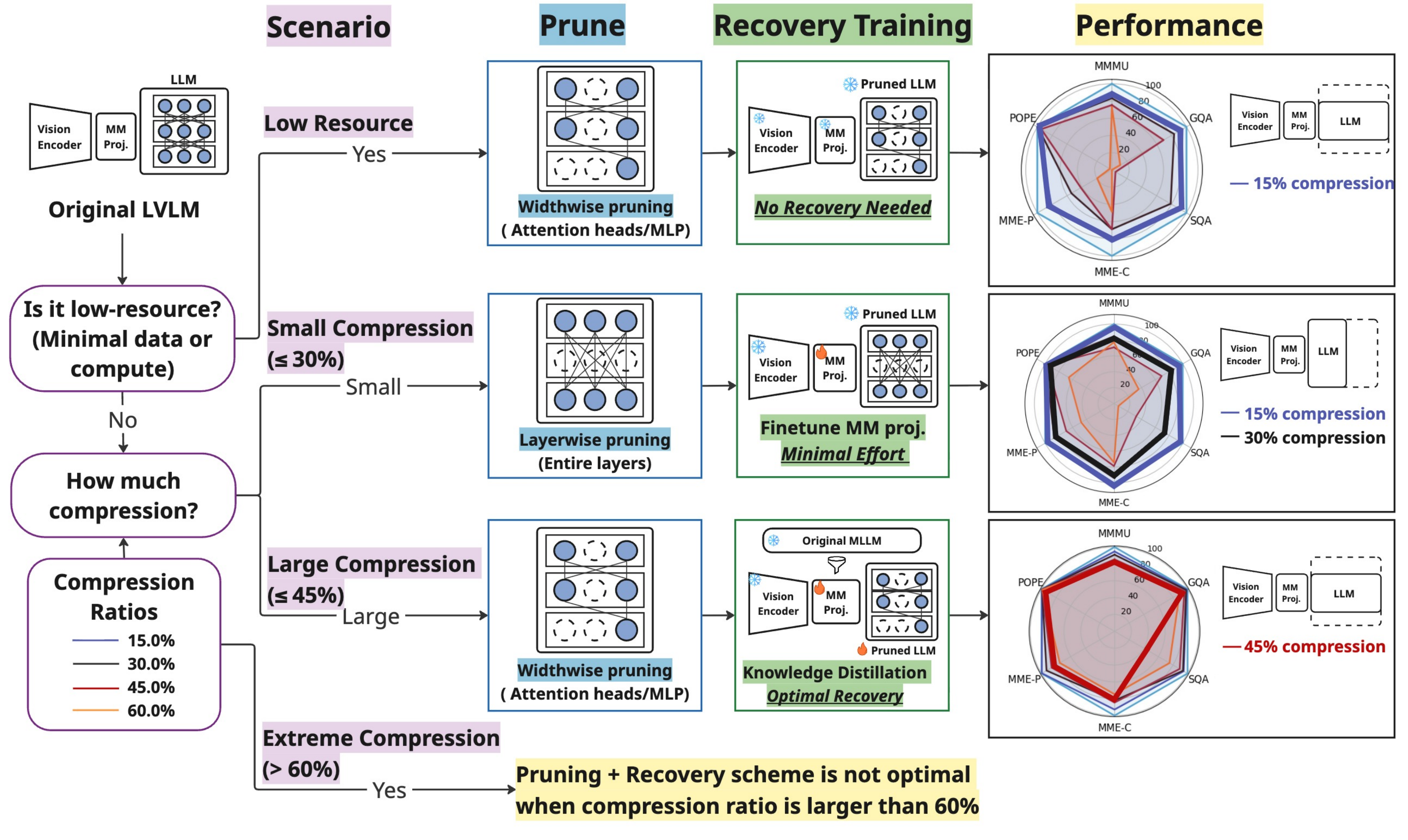}
    \caption{Compression decision flow for LVLMs. The left panel presents a decision flowchart that guides the choice of \textcolor{prune}{\textbf{pruning}} and \textcolor{recovery}{\textbf{recovery}} based on the given \textbf{\textcolor{scenario}{scenarios}}, i.e., resource availability and compression ratio requirements. (i) widthwise pruning only (no recovery) in \textit{extremely low‐resource settings}; (ii) layerwise pruning with MM‐projector fine‐tuning for \textit{moderate compression} ($\leq 30\%$); and (iii) widthwise pruning + knowledge distillation for \textit{high compression} ($\le 45\%$). The right panel shows spider plots of the retained \textbf{\textcolor{performance}{performance} }across multimodal benchmarks, demonstrating each strategy's effectiveness at various compression levels}
    \label{fig:spiderweb_plot}
\end{figure*}
State-of-the-art LVLMs~\cite{chu2023mobilevlm,chen2024internvl, liu2024improved} based on Large Language Models (LLMs)~\cite{touvron2023llama,jiang2024identifying} require substantial resources. 
For instance, models in the LLaVA~\cite{liu2024improved} family typically range from 7 to 34 billion parameters, and even compact models like Bunny-v1.0 (3 billion parameters)~\cite{he2024efficient} pose significant deployment challenges in resource-constrained environments.
Reducing the size of these models without compromising performance is crucial for adapting them to diverse deployment scenarios with varying resource constraints.

Since the language backbone typically constitutes the vast majority of an LVLM's parameters, it represents the primary target for compression~\cite{liu2024improved,he2024efficient,chen2024internvl}. Existing approaches address this mainly by building LVLMs from Small Language Models (SLMs)~\cite{zhu2024comprehensive,he2024efficient,chu2023mobilevlm}. However, these methods suffer from fundamental limitations: they are constrained by the fixed size of the underlying SLM and require expensive, from-scratch training to meet target specifications. In contrast, we investigate an orthogonal and more flexible approach: directly compressing existing LVLMs via structured pruning of their language backbone. This strategy allows customizable model sizing and performance tuning without retraining from scratch.

However, simply applying established pruning techniques to this new domain is non-trivial. While pruning has been extensively studied for unimodal LLMs, its effectiveness in LVLMs remains largely unexplored. LVLMs present unique challenges: pruning the language backbone risks not only degrading linguistic capabilities but also disrupting the learned vision-language alignment. Consequently, understanding how different pruning paradigms and recovery strategies interact under varying resource constraints is essential for practitioners, yet no systematic study currently exists.

To address this gap, we conduct a comprehensive investigation into structured pruning for LVLMs. We focus on structured pruning because it provides immediate efficiency gains on commodity hardware by physically reducing matrix dimensions, unlike unstructured pruning, which requires specialized sparse kernels to realize speedups~\cite{sunsimple,frantar2023sparsegpt}. 
Concretely, we apply two pruning paradigms originally developed for LLMs to the LVLM setting. The first, layerwise pruning~\cite{men2024shortgpt}, removes entire transformer blocks, leveraging evidence that many layers are redundant. The second, widthwise pruning~\cite{ma2023llm}, drops unimportant attention heads and MLP neurons, reflecting observations that only a subset of these sub-components is essential. Crucially, we pair these pruning strategies with various recovery training methods, including supervised finetuning and knowledge distillation on both logits and hidden states. Finally, we vary the pruning ratio and the amount of available data to map out the accuracy/efficiency frontier.

This work extends our preliminary conference version~\cite{huang2025investigating} significantly in scope and depth. First, we expand the empirical evaluation from two to three representative LVLM families by adding Mini-InternVL-Chat-4B~\cite{chen2024internvl}, in addition to LLaVA-v1.5-7B~\cite{liu2024improved} and Bunny-v1-3B~\cite{he2024efficient}, ensuring our findings generalize across different architectures. Second, we broaden the evaluation suite to include domain-specific and reasoning-intensive benchmarks such as AI2D~\cite{kembhavi2016diagram}, MathVista~\cite{lu2023mathvista}, and DocVQA~\cite{mathew2021docvqa}, providing a more granular view of how compression affects complex reasoning versus simple perception. Third, we introduce a novel analysis disentangling the degradation of modality alignment versus language capabilities, offering an explanation for performance drops. Finally, we provide a new sensitivity analysis regarding calibration dataset size and a qualitative study of failure modes, offering a comprehensive guide for practical deployment.
Our systematic empirical analysis provides insights into how pruning and recovery techniques affect LVLM performance under varying compression levels and data availability conditions. 
Specifically, we found:
\begin{itemize}
        \item We confirm that widthwise pruning outperforms layerwise pruning in low-resource settings. A new analysis of reasoning benchmarks reveals that layerwise pruning disproportionately degrades performance on complex reasoning tasks compared to widthwise pruning.
        \item Degradation Dynamics: We demonstrate that low compression ratios primarily damage multimodal alignment rather than the language backbone, whereas high ratios degrade both. This finding explains why finetuning only the projector is sufficient for mild compression.
        \item Optimal Recovery: We validate that Supervised Finetuning (SFT) combined with hidden-state distillation (L2) consistently yields the best recovery.
        \item Data Efficiency and Robustness: We show that calibration is highly data-efficient. Furthermore, for recovery training, effective results can be achieved with as little as 5\% of the original training data.
\end{itemize}
We highlight our key findings in Fig.~\ref{fig:spiderweb_plot}. 
Our findings enable practitioners to efficiently compress LVLMs, allowing researchers to build upon empirically supported strategies without undertaking extensive experimentation themselves. 

\section{Related Work}
\label{related_work}

\subsection{Pruning.} Model compression via pruning has been extensively studied to alleviate computational burdens.
Early approaches focused on \emph{unstructured pruning}~\cite{dong2017learningprunedeepneural,frankle2019lotterytickethypothesisfinding,lee2020signalpropagationperspectivepruning,park2020lookaheadfarsightedalternativemagnitudebased,sanh2020movementpruningadaptivesparsity,farina2024multiflow}, which zeroes out individual weights. While these methods achieve high sparsity with minimal accuracy loss, they typically require specialized hardware or sparse kernels for realizing actual speedups.
\emph{Semi-structured pruning} offers a middle ground~\cite{zhoulearning}, though it often imposes specific hardware constraints.
Consequently, \emph{structured pruning}~\cite{hoefler2021sparsity,li2017pruning,liu2021groupfisherpruningpractical,you2019gatedecoratorglobalfilter,zhang2022all} has gained traction for its ability to remove coherent parameter groups, directly reducing model size and latency on general-purpose hardware.

In the context of Large Language Models (LLMs), recent work demonstrates that structured pruning can be effectively applied to different architectural dimensions. This includes \emph{layerwise} pruning, which removes full layers~\cite{chenstreamlining,songsleb,dery2024everybodyprunenowstructured, men2024shortgpt}, and \emph{widthwise} pruning, which targets components such as attention heads and MLP neurons~\cite{fang2023depgraph,ma2023llm,xia2024shearedllamaacceleratinglanguage}, often with only sparse performance degradation.
Our study builds on these advances by applying both layerwise and widthwise structured pruning to the language backbone of LVLMs, systematically evaluating their efficacy when paired with subsequent recovery training.

% distillation
\subsection{Knowledge Distillation (KD).} KD facilitates the transfer of learned patterns from a high-capacity ``teacher'' model to a more compact ``student'' model~\cite{hinton2015distillingknowledgeneuralnetwork, Gou_2021, sanh2020distilbertdistilledversionbert}. 

In the context of language modeling, this transfer is typically achieved by aligning the student’s output distributions with those of the teacher~\cite{liang2021mixkdefficientdistillationlargescale, hsieh2023distillingstepbystepoutperforminglarger}. Beyond final outputs, the student can also be trained to mimic the teacher's internal representations, such as hidden states~\cite{jiao2020tinybertdistillingbertnatural, sun2019patientknowledgedistillationbert} or self-attention patterns~\cite{wang2020minilmdeepselfattentiondistillation, wang2021minilmv2multiheadselfattentionrelation}.
Recent methodological refinements have further optimized this process. For instance, multi-stage distillation allows for the transfer of intermediate features to capture more granular model behaviors~\cite{hsieh2023distillingstepbystepoutperforminglarger}, while~\cite{gu2023knowledge} utilizes reverse Kullback-Leibler (KL) divergence to prevent the student from over-focusing on the teacher’s low-probability output regions. Notably, KD has emerged as a recovery strategy following aggressive pruning~\cite{hoffmann2021towards, muralidharan2024compact}, in which the uncompressed model serves as the teacher to restore lost performance.

While KD is increasingly used in the vision-language domain to distill large LVLMs into smaller architectures~\cite{cai2025llava,kim2025compodistill}, its use as a \textit{post-pruning recovery mechanism} for LVLMs remains under-explored. We address this gap by empirically evaluating various KD strategies against conventional supervised finetuning to identify the most effective recovery schemes for pruned LVLMs.

% Multimodal Compression
\subsection{Efficient LVLMs.} 
Research on accelerating LVLMs generally follows two paradigms: input-level optimization and architectural efficiency.

A significant body of work focuses on visual token reduction~\cite{shang2025llava,tang2025covipal,wang2025autoprune}. For instance, LLaVA-PruMerge~\cite{shang2025llava} utilizes a dynamic approach of token pruning and merging for adaptive compression, while CoViPAL~\cite{tang2025covipal} introduces a model-agnostic plug-and-play module for token pruning. While these token-level methods reduce computational overhead by lowering input dimensionality, they differ fundamentally from our approach as they leave the underlying model architecture uncompressed.

Alternatively, researchers have developed lightweight LVLMs by integrating smaller language backbones, as seen in LLaVA-Phi~\cite{zhu2024llava}, MobileVLM~\cite{chu2023mobilevlm}, and Bunny~\cite{he2024efficient}. While these models are inherently efficient, they are constrained by the fixed dimensions of their underlying small language models (SLMs). In contrast, our study focuses on the structured pruning and recovery of existing, large-scale LVLMs. This enables customizable model sizing and performance tuning, offering greater flexibility than training small-scale models from scratch.

\section{Methodology}
\label{Methodology}

This section outlines our approach to compressing LVLMs. We first introduce two pruning strategies: layerwise and widthwise pruning. We then describe methods to recover model performance through supervised finetuning and knowledge distillation.

\xhdr{Notation.} Given a triplet $\mathbf{X} = \{\mathbf{x}_v, \mathbf{x}_p, \mathbf{x}_r\}$, the objective of an LVLM $m_\theta$, parameterized by $\theta=\{\psi,\phi,\mathbf{W}\}$, is to generate a response $\mathbf{x}_r$ based on an input image $\mathbf{x}_v$ and a text prompt $\mathbf{x}_p$, such that $m_\theta(\mathbf{x}_v, \mathbf{x}_p) = \mathbf{x}_r$. The LVLM typically consists of a vision encoder $g_\psi(\cdot)$, an LLM $f_{\phi}(\cdot)$, and a multimodal projector $\mathbf{W}$ aligning the two modalities. The prompt $\mathbf{x}_p$ is tokenized into $\mathbf{T}_p$, while the vision encoder processes the image $\mathbf{x}_v$ to extract visual features, which are then converted into language embedding tokens $\mathbf{T}_v$ via the multimodal projector:
\begin{equation}
    \mathbf{T}_v=\mathbf{W} \cdot g_\psi(\mathbf{x}_v) \ \ \text{and} \ \ f_\phi(\mathbf{T}_v \odot \mathbf{T}_p) = \mathbf{x}_r.
\end{equation}

The concatenated visual tokens $\mathbf{T}_v$ and prompt tokens $\mathbf{T}_p$ are fed into the LLM’s $M$ layers, producing hidden states $\{{\mathbf{H}_i} \in \mathbb{R}^{T \times d}\}^M_{i=1}$, where $T$ is the number of tokens and $d$ is the hidden dimension. Finally, the probabilities $p_{m_\theta}(\mathbf{x}_r|\mathbf{x}_v, \mathbf{x}_p,\tau)$ are computed by passing the final hidden state through the classification head with softmax temperature $\tau$.

\subsection{Pruning.}
Large Transformers are largely over-parameterized, as whole layers can be dropped with little accuracy loss~\cite{fan2019reducing,sajjad2023effect}, and only a few attention heads or MLP units per layer truly matter~\cite{voita2019analyzing,michel2019sixteen,mccarley2019structured,hudson2019gqa}.
Motivated by these findings, we explore two pruning paradigms specifically targeting the language model backbone within LVLMs: layerwise pruning, which removes entire transformer layers, and widthwise pruning, which eliminates the least important components within each layer.
To determine which layers or components to prune, we draw a small subset of $n$ samples from the original visual instruct-tuning dataset as the calibration dataset $\mathcal{D} = \{\mathbf{x}^j_v, \mathbf{x}^j_p, \mathbf{x}^j_r\}_{j=1}^{n}$. The importance of each layer or component is assessed, and the lowest-importance components are pruned.

\xhdr{Layerwise Pruning.}
\label{layer_prune}
To identify the redundant layers, we use the Block Influence (BI) score~\cite{men2024shortgpt}, which quantifies the importance of layer $i$ through the cosine distance between input $\mathbf{H}_i$ and output hidden states $\mathbf{H}_{i+1}$. The key assumption is that layers that cause larger changes in hidden states have a greater influence on model performance. The BI score of layer $i$ is then calculated by
\begin{equation}
    \text{BI}_i(\mathcal{D}) = 1 - \mathbb{E}_{\mathbf{X} \sim \mathcal{D}, t}\left[\frac{\mathbf{H}^\mathsf{T}_{i,t} \mathbf{H}_{i+1,t}}{\|\mathbf{H}_{i,t}\|_2 \|\mathbf{H}_{i+1,t}\|_2}\right],
\end{equation}
where $\mathbf{H}_{i,t}$ represents the $t^{th}$ row of $\mathbf{H}_i$. After calculating the BI scores, the layers are ranked by importance, and the lowest-scoring layers are pruned. 

\xhdr{Widthwise Pruning.}
\label{width_prune}
To address the widthwise redundancy, we apply dependency-based structural pruning. Following~\cite{fang2023depgraph} and~\cite{ma2023llm}, we build a dependency graph inside each LLM layer. Let $N_{i}$ and $N_{j}$ represent two neurons in the layer, where $\text{In}(N_{i})$ and $\text{Out}(N_{i})$ represent the neurons connected to $N_{i}$ as inputs and outputs, respectively. Neuron $N_{j}$ is dependent on on $N_{i}$ if
\begin{equation}
\begin{split}
    N_{j} \in \text{Out}(N_{i}) \land  \text{Num}_{\text{In}(N_{j})} = 1, \\
    \text{or } N_{j} \in \text{In}(N_{i}) \land  \text{Num}_{\text{Out}(N_{j})} = 1
\end{split}
\end{equation}
where $\text{Num}_{\text{In}(N_{j})}$ refers to the number of input neurons of $N_{j}$ and $\text{Num}_{\text{Out}(N_{j})}$ is the number of the output neurons of $N_{j}$. In words, $N_{i}$ is the only downstream or upstream node of $N_{j}$. If neuron $N_{i}$ is pruned, all its dependent neurons $N_{j}$ must also be pruned. This process results in a set of dependency graphs ${{G}} = \{w_i^k\}_{i=1}^M$, where $M$ is the number of structures in the graph and ${w_i^k}$ represents the $k^{th}$ weight parameter within a structure. 

We assess their importance at the group level since all weights within a graph must be pruned together. Group importance is evaluated by comparing the loss of vision language modeling ${\displaystyle \mathcal{L}_{CE}(m_\theta(\mathbf{x}_v,\mathbf{x}_q),\mathbf{x}_r)}$ in the calibration data set, with and without weight. To efficiently approximate the importance, we apply a Taylor expansion using gradient information:
\begin{equation}
\begin{split}
    I_{w_i^k}(\mathbf{X}) &= |\mathcal{L}_{CE}(\mathbf{X},m_\theta) - \mathcal{L}_{CE}(\mathbf{X},m^{w_i^k=0}_\theta)| \\
    &\approx \left| \frac{\partial \mathcal{L}_{CE}(\mathbf{X}, m_\theta)}{\partial w_i^k} w_i^k \right| \,.
\end{split}
\end{equation}

We then prune the graphs with the lowest group importance $I_{G}$:
 \begin{equation}
     \textstyle{I_{G}(\mathcal{D}) = \mathbb{E}_{\mathbf{X}\sim \mathcal{D}} \left[\sum_{i}^M\sum_{k}{I_{w_i^k}}(\mathbf{X})\right]} \,.
 \end{equation}

\subsection{Recovery Training.}
\label{Retraining}
Pruning LVLMs degrades performance, affecting both language modeling and cross-modality alignment. To mitigate this, we investigate two recovery training methods: supervised finetuning (Sec. \ref{Retraining_sft}) and knowledge distillation (Sec. \ref{retraining_kd}). We consider the original uncompressed model as the teacher $m^\text{T}_\theta$, the pruned model as the student $m^\text{S}_{\theta^\prime}$, and a recovery dataset $\mathcal{D}$.

\subsubsection{Recovery Training with Supervised Finetuning~(SFT).}
\label{Retraining_sft}
% A simple yet effective approach to recovery training is supervised finetuning on the original dataset. 
% This method helps counteract performance degradation by allowing the model to adapt its parameters to the modified architecture while taking advantage of the detailed annotations in the original dataset. 
We first focus on training only the multimodal projector to realign the vision and language spaces.
Second, we jointly finetune the projector and the pruned language model while keeping the vision encoder fixed, as finetuning the vision encoder has been shown to yield negligible gains~\cite{karamcheti2024prismatic} and increases training cost.
We use the cross-entropy loss for supervised finetuning:
\begin{equation}
    \mathcal{L}_{sft}(m^\text{S}_{\theta^\prime}, \mathcal{D})=\mathbb{E}_{\mathbf{X}\sim\mathcal{D}}[\mathcal{L}_{CE}(m^\text{S}_{\theta^\prime}(\mathbf{x}_v,\mathbf{x}_p),\mathbf{x}_r)].
\end{equation}

\subsubsection{Recovery Training with Knowledge Distillation.}
\label{retraining_kd}
KD
% is a method used to transfer knowledge from a large, well-trained model (the teacher) to a smaller or pruned model (the student) \cite{hinton2015distillingknowledgeneuralnetwork}. This approach 
allows the pruned model to regain lost performance by mimicking the decision-making process of the more capable teacher (original model). 
% In our setup, the uncompressed model acts as the teacher, while the pruned model serves as the student. 
We explore two main strategies, logits-based KD and hidden state based KD.
% , evaluating different loss functions and their trade-offs.

\xhdr{Logits-based KD} aligns the output probability distributions of the pruned model with those of the teacher model. The logits-based KD loss is defined as
\begin{equation}
\begin{split}
    \mathcal{L}_{logits}(m^\text{S}_{\theta^\prime}, &m^\text{T}_\theta, \mathcal{D}) = \\
    \mathbb{E}_{\mathbf{X}\sim\mathcal{D}} \big[ & \mathcal{L}_{KD}(p_{m^\text{T}_\theta}(\mathbf{x}_r|\mathbf{x}_v, \mathbf{x}_p,\tau), \\
    & p_{m^\text{S}_{\theta^\prime}}(\mathbf{x}_r|\mathbf{x}_v, \mathbf{x}_p,\tau)) \big].
\end{split}
\end{equation}

We explore two losses to evaluate the differences between the logit distributions of the student $p_\theta$ and the teacher $q_{\theta^\prime}$: Kullback–Leibler divergence (KL), denoted as  $\mathbf{KL} (p_\theta \Vert q_{\theta^\prime})$ and its reversed form (RKL), denoted as ${\mathbf{KL}(q_{\theta^\prime} \Vert p_\theta)}$. The standard KD objective, minimizing the approximated forward KL, encourages the student distribution to match all modes of the teacher distribution.
In contrast, using RKL encourages $q_{\theta^\prime}$ to focus on the major modes of $p_\theta$ while assigning low probabilities to its less significant regions. This helps the student model avoid learning unnecessary long-tail variations of the teacher distribution and instead focus on generating more accurate responses~\cite{gu2023knowledge,holtzman2019curious}.

\xhdr{Hidden State Matching} involves aligning the pruned model’s intermediate representations (hidden states) $\mathbf{H}^{m^\text{S}_{\theta^\prime}}_i$ with the teacher model's $\mathbf{H}^{m^\text{T}_\theta}_i$. The corresponding loss for a layer $i$ can be defined as
\begin{equation}
    \mathcal{L}_{match}(m^\text{S}_{\theta^\prime}, m^\text{T}_\theta, \mathcal{D}) = \mathbb{E}_{\mathbf{X}\sim\mathcal{D}}\left[\mathcal{L}_{feat}(\mathbf{H}^{m^\text{S}_{\theta^\prime}}_i,\mathbf{H}^{m^\text{T}_\theta}_i)\right],
\end{equation}
where $\mathcal{L}_{feat}$ refers to a feature matching loss. Both~\cite{yang2024clip} and~\cite{popp2024zero} suggest that applying a feature-based L2 distillation loss improves the student model’s performance, particularly for pre-trained vision-language models. Consequently, we employ L2 loss as the feature matching loss $\mathcal{L}_{feat} (x, y)=\|x -y\|_2^2$. 
The total loss for recovery training is computed as:
\begin{equation}
\begin{split}
    \mathcal{L}(m^\text{S}_\theta, m^\text{T}_\theta, \mathcal{D}) &=  \alpha \mathcal{L}_{sft}(m^\text{S}_{\theta^\prime}, \mathcal{D}) \\&+ \beta \mathcal{L}_{logits}(m^\text{S}_{\theta^\prime}, m^\text{T}_\theta, \mathcal{D}) \\ & + \gamma \mathcal{L}_{match}(m^\text{S}_{\theta^\prime}, m^\text{T}_\theta, \mathcal{D})
\end{split}
\end{equation}
where $\alpha$, $\beta$, and $\gamma$ are the coefficients that balance three loss components.

\section{Experiments}
\label{Experiments}
 In this section, we first introduce our experimental setup and then demonstrate the main findings on model pruning (Sec. \ref{Experiments_pruning}) and performance recovery (Sec. \ref{Supervised Finetuning for Performance Recovery After Pruning}, Sec. \ref{kd_for_recovery}). We highlight the findings on recovery training using only a small fraction of data (Sec. \ref{Experiments_few_data}) and present the model compression results following our best practices (Sec. \ref{Model compression results following our best practices}), including the qualitative performance and failure case. 
% Additionally, we provide qualitative insights into the compressed models (Sec. \ref{Qualitative results of the compressed models}). Next, we compare and integrate our best practices with quantization techniques (Sec. \ref{Pruning vs. Quantization}). 

% \input{tables/architecture_table}
\begin{table*}[h]
\caption{Architecture details of the uncompressed models. We present the number of parameters, along with the vision encoder, multimodal projector and the language decoder of the models included in our study}
\centering
\resizebox{\linewidth}{!}{
\begin{tabular}{ccccc}
\toprule
\multicolumn{1}{c}{\bf Model}
&\multicolumn{1}{c}{\bf Parameters}
&\multicolumn{1}{c}{\bf Vision Encoder}
&\multicolumn{1}{c}{\bf Multimodal Projector}
&\multicolumn{1}{c}{\bf Language Decoder}
\\ \midrule 
LLaVA-v1.5-7B & 7.0B & CLIP-ViT-L~(0.3B) & mlp2x-gelu~(0.01B) & Vicuna-v1.5~(6.7B) \\
Mini-InternVL-Chat-4B-V1.5 & 4.1B &  InternViT-300M-448px L~(0.3B) & mlp2x-gelu~(0.02B) & Phi-3-mini-128k-instruct~(3.8B) \\
Bunny-v1.0-3B & 3.2B & SigLIP-SO~(0.4B) & mlp2x-gelu~(0.02B) & Phi-2~(2.8B)
\\ \bottomrule
\end{tabular}
}
% }
\label{model architecture}
\end{table*}
\subsection{Experimental Setup.} 
\xhdr{Model.} To ensure architectural diversity, we evaluate pruning and recovery methods on three LVLMs from distinct model families: LLaVA-v1.5-7B (LLaVA)~\cite{liu2024improved}, Mini-InternVL-Chat-4B (InternVL)~\cite{chen2024internvl}, and Bunny-v1.0-3B (Bunny)~\cite{he2024efficient}. As detailed in Table~\ref{model architecture}, the language backbone accounts for the majority of the total parameter count across all three architectures. This structural imbalance underscores the need to target the language component to achieve meaningful compression.

\xhdr{Data.} We exclusively utilize each model's original visual instruction tuning dataset for both pruning calibration and recovery training: LLaVA-v1.5-mix665k~\cite{liu2024improved} for LLaVA, InternVL-Chat-V1-2-SFT-Data~\cite{chen2024internvl} for InternVL, and Bunny-695K~\cite{he2024efficient} for Bunny.
During the pruning stage, we draw a small subset of $n$ samples from the respective training data to serve as the calibration dataset for computing importance scores.
% \new{We provide additional analysis on the impact of calibration dataset size.}
For the subsequent recovery phase, we investigate the impact of data scale by finetuning on varying fractions (5\%, 10\%, 20\%, and 100\%) of the full dataset. 

% During pruning, we randomly sample data from the training dataset as the calibration dataset to compute the importance. For recovery training, we experiment with various portions of the original dataset ~( 5\%, 10\%, 20\%, and 100\%) from the original visual instruction tuning dataset for recovery. 
% \new{
% We set the distillation temperature to 2.0 for logits-based distillation and use the final layer representation for hidden state matching}

\xhdr{Evaluation.}
We evaluate the pruned and recovered models on a diverse suite of benchmarks designed to challenge various facets of multimodal capability. To assess domain-specific knowledge, we employ MMMU~\cite{yue2023mmmu}, SQA~\cite{lu2022learn}. For advanced reasoning, we employ MathVista~\cite{lu2023mathvista} and AI2D~\cite{kembhavi2016diagram}. 
Complementing these are assessments of fine-grained perception and hallucination, including POPE~\cite{li2023evaluating} and the separate Cognition and Perception subsets of MME~\cite{yin2023survey}. Finally, we evaluate general visual understanding and document understanding via GQA~\cite{hudson2019gqa} and DocVQA~\cite{mathew2021docvqa}. All assessments are conducted using the \texttt{lmms-eval} suite~\cite{lmms_eval2024} to ensure reproducibility. 
Crucially, these benchmarks span a wide range of prompt formats, requiring the models to adapt to multiple-choice questions (e.g., MMMU, SQA, AI2D), binary Yes/No classification (e.g., POPE, MME), and open-ended free generation (e.g., MathVista, DocVQA, GQA). For clarity, we report performance as a percentage relative to the uncompressed model’s score on each benchmark.

\subsection{Structural Pruning Paradigms: Widthwise vs. Layerwise Analysis.}
\label{Experiments_pruning}
In this section, we compare two structural pruning strategies: widthwise pruning and layerwise pruning. We evaluate the relative strengths of these methods by analyzing their zero-shot performance immediately after pruning in Sec. \ref{Zero-Shot Performance.}, their post-training recovery capabilities after finetuning in Sec. \ref{The Impact of Recovery Training.}, and their efficiency regarding actual GPU memory and computational cost reduction in Sec. \ref{Hardware Efficiency.}.
\begin{table*}[t]
\caption{Pruning results for LLaVA-v1.5-7B, Mini-InternVL-Chat-4, and Bunny-v1-3B across benchmarks. Widthwise pruning generally results in better performance without finetuning compared to layerwise pruning}\label{tab:pruning_results}
\begin{center}
\resizebox{\linewidth}{!}
{
\begin{tabular}{lrrrrrrrrrrrrr}
\toprule
\textbf{Method} & \textbf{Size} & \textbf{Ratio} & \textbf{MMMU} & \textbf{SQA} & \textbf{MathVista} & \textbf{AI2D} & \textbf{MME-C} & \textbf{MME-P} & \textbf{POPE} & \textbf{DocVQA} & \textbf{GQA} & \textbf{AVG} & \textbf{AVG-\%} \\
\midrule
LLaVA-v1.5-7B & 7.0B & 0\% & 35.10 & 68.67 & 26.70 & 54.80 & 363.21 & 1511.33 & 86.99 & 28.10 & 61.98 & 53.70 & 100.00\% \\
\midrule
Widthwise & 6.0B & 15\% & 32.40 & 63.21 & 23.10 & 45.40 & 268.93 & 1432.47 & 86.57 & 23.90 & 59.34 & 48.80 & 90.87\% \\
 & 5.0B & 30\% & 31.00 & 54.29 & 15.80 & 37.80 & 253.21 & 1174.93 & 86.29 & 17.90 & 52.59 & 42.90 & 79.88\% \\
 & 3.8B & 45\% & 27.60 & 12.10 & 5.40 & 13.40 & 70.00 & 347.45 & 45.96 & 5.30 & 20.86 & 17.42 & 32.43\% \\
 & 2.8B & 60\% & 23.30 & 0.40 & 1.30 & 1.20 & 2.14 & 19.24 & 3.94 & 0.80 & 0.43 & 3.62 & 6.75\% \\
\midrule
Layerwise & 6.0B & 15\% & 32.70 & 59.64 & 14.80 & 32.10 & 210.71 & 921.88 & 78.69 & 16.50 & 42.18 & 38.78 & 72.22\% \\
 & 5.0B & 30\% & 31.80 & 55.23 & 13.50 & 29.70 & 202.14 & 701.83 & 86.38 & 14.60 & 42.77 & 37.15 & 69.18\% \\
 & 4.0B & 45\% & 26.90 & 3.82 & 5.40 & 10.50 & 132.86 & 616.63 & 51.69 & 6.10 & 14.39 & 18.47 & 34.40\% \\
 & 2.8B & 60\% & 25.80 & 0.00 & 0.00 & 0.00 & 0.00 & 0.00 & 0.00 & 0.00 & 0.00 & 2.87 & 5.34\% \\
\midrule
Mini-InternVL-Chat-4B & 4.1B & 0\% & 43.20 & 93.30 & 53.70 & 76.90 & 547.50 & 1596.71 & 88.00 & 87.70 & 62.57 & 72.63 & 100.00\% \\
\midrule
Widthwise & 3.5B & 15\% & 42.70 & 92.96 & 38.80 & 72.40 & 527.86 & 1594.37 & 88.09 & 80.20 & 58.43 & 68.81 & 94.74\% \\
 & 3.0B & 30\% & 33.30 & 68.82 & 28.80 & 44.50 & 397.14 & 1300.60 & 85.20 & 64.20 & 42.39 & 53.54 & 73.72\% \\
 & 2.5B & 45\% & 24.00 & 12.30 & 6.20 & 10.80 & 260.20 & 980.00 & 83.00 & 25.00 & 22.00 & 29.43 & 40.52\% \\
 & 1.6B & 60\% & 22.30 & 0.00 & 0.40 & 0.50 & 0.51 & 14.34 & 10.00 & 0.00 & 3.72 & 5.12 & 7.05\% \\
\midrule
Layerwise & 3.5B & 15\% & 43.60 & 93.12 & 22.50 & 70.20 & 510.10 & 1588.30 & 88.15 & 78.30 & 52.35 & 65.71 & 90.48\% \\
 & 3.0B & 30\% & 28.30 & 62.82 & 18.40 & 24.50 & 220.90 & 1200.60 & 73.20 & 53.10 & 27.39 & 41.71 & 57.42\% \\
 & 2.5B & 45\% & 24.68 & 8.80 & 1.20 & 8.80 & 220.10 & 887.35 & 80.00 & 12.00 & 16.00 & 24.82 & 34.17\% \\
 & 1.6B & 60\% & 21.30 & 0.00 & 0.40 & 0.50 & 0.31 & 6.34 & 0.20 & 0.00 & 2.53 & 3.48 & 4.79\% \\
\midrule
Bunny-v1-3B & 3.2B & 0\% & 34.10 & 70.70 & 25.80 & 60.49 & 289.30 & 1487.71 & 87.82 & 22.80 & 54.72 & 51.89 & 100.00\% \\
\midrule
Widthwise & 2.8B & 15\% & 30.90 & 65.64 & 23.40 & 55.73 & 242.50 & 1207.85 & 87.94 & 18.20 & 51.83 & 47.15 & 90.87\% \\
 & 2.4B & 30\% & 28.40 & 55.73 & 14.20 & 43.97 & 199.64 & 807.95 & 87.13 & 12.50 & 45.65 & 39.21 & 75.58\% \\
 & 2.0B & 45\% & 25.70 & 3.42 & 5.10 & 5.56 & 200.00 & 618.25 & 83.12 & 4.80 & 37.92 & 24.61 & 47.44\% \\
 & 1.5B & 60\% & 24.80 & 0.00 & 1.20 & 0.20 & 141.07 & 293.23 & 2.34 & 0.80 & 6.12 & 7.53 & 14.51\% \\
\midrule
Layerwise & 2.8B & 15\% & 33.80 & 69.66 & 18.80 & 52.73 & 271.43 & 1456.41 & 87.91 & 15.80 & 29.42 & 46.10 & 88.84\% \\
 & 2.4B & 30\% & 29.00 & 28.76 & 10.10 & 34.97 & 272.86 & 1273.34 & 86.50 & 10.50 & 24.77 & 35.82 & 69.03\% \\
 & 2.0B & 45\% & 23.90 & 3.47 & 8.40 & 3.56 & 191.43 & 867.37 & 80.09 & 11.20 & 16.85 & 23.86 & 45.99\% \\
 & 1.5B & 60\% & 26.60 & 17.15 & 4.20 & 0.20 & 0.71 & 55.92 & 0.02 & 3.90 & 0.02 & 6.11 & 11.77\% \\
\bottomrule
\end{tabular}
}
\end{center}
\end{table*}
\begin{figure*}[t]
    \centering
    \includegraphics[width=\linewidth]{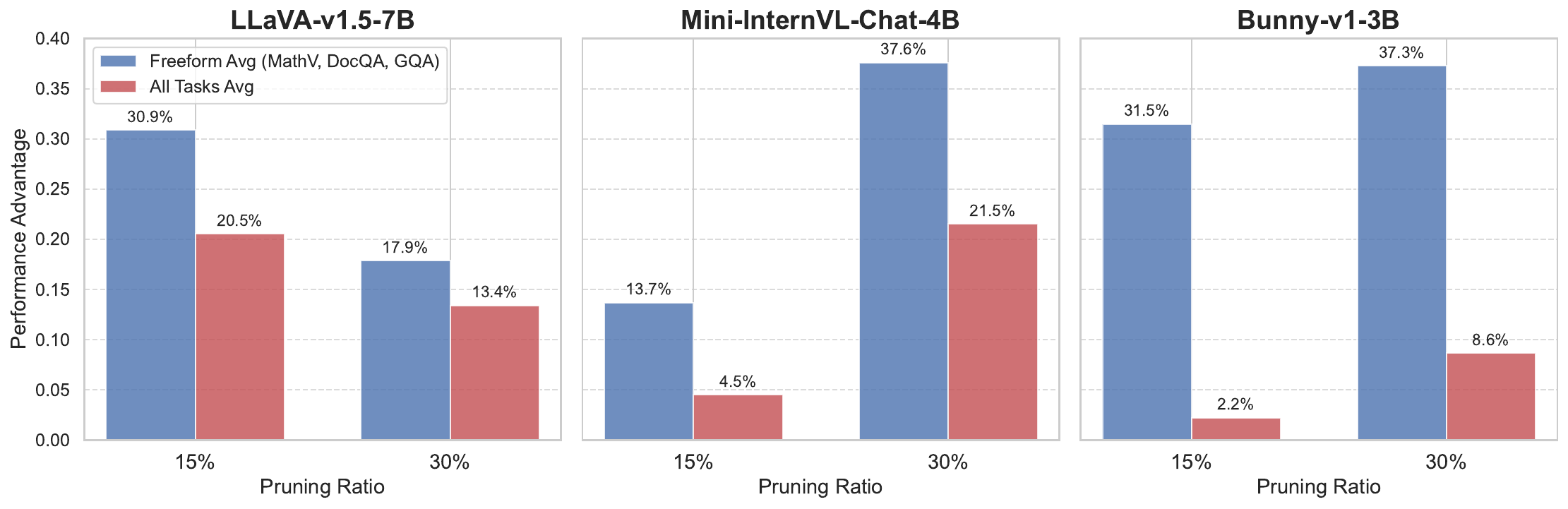}
    \caption{\textbf{Comparative Analysis of widthwise vs. layerwise Pruning.} We evaluate the relative performance advantage of widthwise over layerwise pruning ($\Delta = (S_{\text{width}} - S_{\text{depth}}) / {S_{\text{width}}}$) at 15\% and 30\% sparsity ratios. Widthwise pruning consistently outperforms layerwise strategies, with a significantly larger gap observed in complex freeform generation tasks (MathVista, DocVQA, GQA) compared to the global average, suggesting depth is critical for open-ended reasoning
    }
    \label{fig:pruning_comparison_tasks}
\end{figure*}
\subsubsection{Zero-Shot Performance.}
\label{Zero-Shot Performance.}
Our evaluation shows that widthwise pruning consistently outperforms layerwise pruning when the model is not finetuned. As shown in Table~\ref{tab:pruning_results}, widthwise pruning keeps much more of the original capability at moderate pruning ratios (15\%-30\%). For example, at a 15\% ratio, LLaVA retains 90.87\% of its performance with widthwise pruning, compared to only 72.22\% with layerwise pruning. However, there is a limit: beyond a 45\% ratio, both methods fail, and benchmark scores on SQA and DocVQA drop to near-zero.

Notably, the contrast between widthwise and layerwise pruning becomes much clearer once we stratify benchmarks by either (i) output format or (ii) reasoning demand. 
First, on tasks that require free-form responses (GQA, DocVQA, and MathVista), the gap is substantially larger than the aggregate trend (Fig.~\ref{fig:pruning_comparison_tasks}). 
For instance, on Bunny at a 15\% pruning ratio, widthwise pruning improves overall performance by only 2.2\% relative to layerwise pruning, whereas the margin on generative tasks increases to 31.5\%. 
Second, a similar pattern appears on reasoning-heavy benchmarks (MathVista and AI2D). 
Table~2 shows that widthwise pruning reaches 65.64 on AI2D versus 52.73 for layerwise pruning, corresponding to an absolute gap of 12.91 points (nearly 20\% relative). 
Taken together, these results indicate that \emph{layer removal} is disproportionately harmful on benchmarks that place higher demands on multi-step reasoning and/or free-form generation.

Crucially, our findings in the multimodal domain strongly align with recent discoveries in the unimodal LLM pruning literature, pointing to a fundamental property of transformer architectures. For instance, Sreenivas et al.~\cite{sreenivas2024llm} recently demonstrated a similar phenomenon when pruning LLMs: widthwise pruning proved vastly superior to layerwise pruning, especially on reasoning-intensive benchmarks such as GSM8K \cite{cobbe2021training}. The fact that layer removal disproportionately impairs complex reasoning across both pure language models and vision-language models underscores that this degradation is not merely an artifact of multimodal integration. Instead, it exposes a structural vulnerability inherent to the underlying language backbone itself.

We attribute these trends to two primary structural factors: feature alignment and architectural depth. First, the general performance drop in layerwise pruning stems from feature misalignment. In a pre-trained network, Layer $N+1$ is optimized to process the specific feature distribution output by Layer $N$. Removing a layer forces Layer $N$ to feed directly into Layer $N+2$. This creates a distribution shift that subsequent layers are not calibrated to handle, disrupting the forward pass for all tasks regardless of complexity. Widthwise pruning avoids this shock by maintaining the original layer interfaces, preserving the expected feature flow.
Second, the disproportionate decline in free-form generation and reasoning performance (e.g., MathVista) is directly linked to the loss of processing depth. Although non-linearity exists in both dimensions, depth is specifically required to build sequential reasoning chains~\cite{telgarsky2016benefits,liu2024mobilellm}. Theoretical studies show that complex logic relies on a deep sequence of transformations, which cannot be efficiently approximated by a wider, shallower network. Layerwise pruning explicitly shortens this ''chain of thought'', limiting the model's ability to perform multi-step reasoning. In contrast, widthwise pruning reduces redundancy within layers but preserves the full sequence of steps required for complex tasks.

\subsubsection{The Impact of Recovery Training after Pruning.}
\label{The Impact of Recovery Training.}
While zero-shot performance highlights the immediate structural damage of pruning, recovery training reveals the model's capacity to heal. In this section, we compare the intrinsic recoverability of widthwise versus layerwise structures under standard conditions, explicitly deferring the optimization of specific training recipes, such as parameter selection and distillation objectives, to Sec.~\ref{Supervised Finetuning for Performance Recovery After Pruning}) and Sec.~\ref{kd_for_recovery}).

\begin{figure*}[t]
    \centering
    \includegraphics[width=\linewidth]{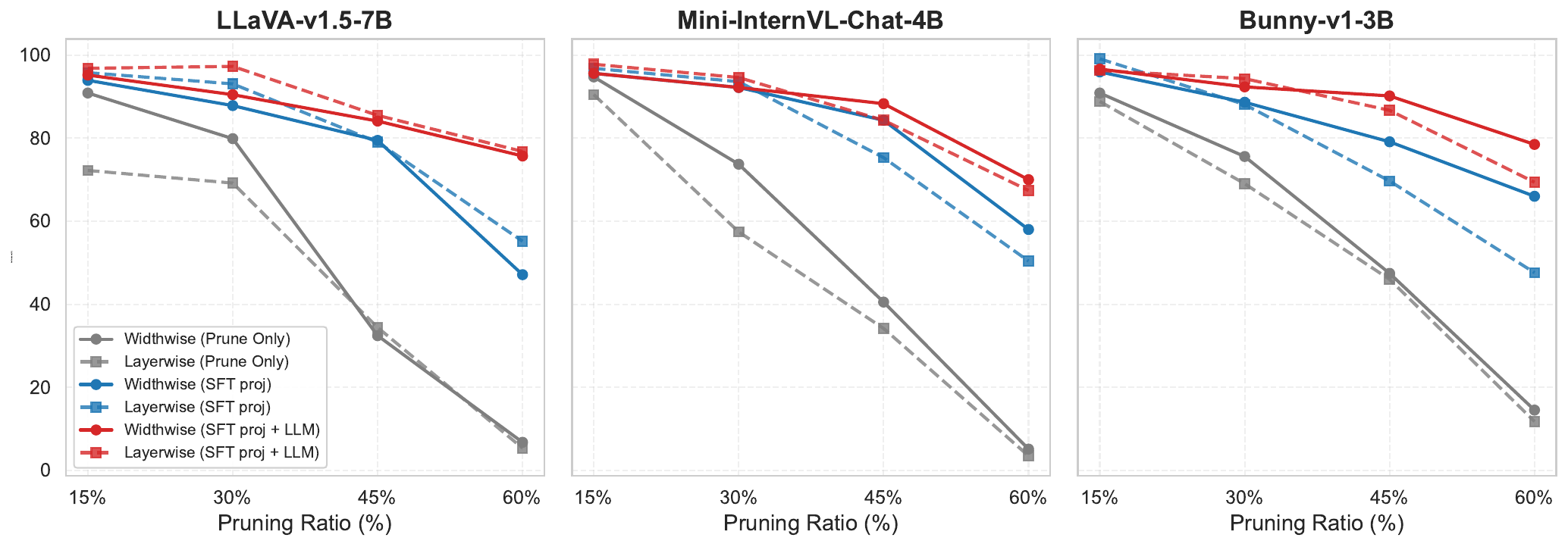}
    \caption{Evolution of pruning efficacy across supervised finetuning(SFT) regimes. The plots illustrate the average performance across LLaVA-v1.5-7B, Mini-InternVL-Chat-4, and Bunny-v1-3B. ((1) Robustness of widthwise pruning: in the absence of SFT (gray lines), widthwise pruning (solid) consistently maintains higher performance than layerwise pruning (dashed). (2) Regime shifts with SFT: Post-SFT (blue/red lines), the landscape inverts at lower ratios ($\le 30\%$), where layerwise pruning becomes superior. However, at higher compression rates ($\ge 45\%$), widthwise pruning regains its advantage. (3) Efficiency of multimodal projector SFT: SFT only the projector (blue) yields performance comparable to full projector+LLM SFT (red) at low sparsity, highlighting a cost-effective strategy for moderate model compression
    }
    \label{fig:finetune_plot}
\end{figure*}

When recovery training is applied, it reshapes the performance landscape (see Fig.~\ref{fig:finetune_plot}, red lines). Layerwise pruning demonstrates superior recoverability at low compression ratios ($\le30\%$). However, at high compression ratios ($\ge45\%$), the trend shifts: widthwise pruning generally yields superior recovery performance across most models evaluated (e.g., Mini-InternVL and Bunny) or, at a minimum, achieves parity with layerwise pruning (as observed with LLaVA).

This reversal is driven by the shifting balance between the two factors identified in Sec.~\ref{Zero-Shot Performance.}: feature misalignment, which is learnable, and processing depth, which is structural. 
At low compression ratios, the primary limitation of layerwise pruning was the immediate feature distribution shift between unconnected layers. Recovery training effectively mitigates this issue by realigning the feature spaces of adjacent layers. Once this misalignment is corrected, layerwise pruning proves advantageous because it preserves the full dimensionality (width) of the remaining layers, maintaining a richer representation subspace than the thinned-out layers of widthwise pruning.
However, as the compression ratio increases, the model hits the structural depth constraint discussed in Sec.~\ref{Zero-Shot Performance.}. While finetuning can address feature alignment, it cannot overcome the fundamental need for sequential non-linear transformations required for complex reasoning~\cite{telgarsky2016benefits}. Thus, widthwise pruning prevails in the high-compression regime simply because it sustains the necessary architectural framework for reasoning, a structural advantage that recovery training alone cannot replicate.

\textbf{\textit{\textcolor{blue!50!black!80}{Takeaway.}}}
Widthwise pruning proves more effective than layerwise pruning in obtaining the best pruned model.  
With recovery training, layerwise pruning shows a slight advantage at compression ratios no greater 30\%, while widthwise pruning performs better at higher compression ratios. 

\subsubsection{Hardware Efficiency.}
\label{Hardware Efficiency.}
Table \ref{tab:model_size_combined} compares the resource efficiency of widthwise and layerwise pruning across three metrics: GPU memory usage, computational cost (FLOPs), and real-world inference latency. As expected, increasing the pruning ratio linearly reduces memory consumption and computational overhead across all models. For instance, at a 60\% compression ratio, resource consumption decreases by approximately half for both methods.

Our results highlight a strong consistency between widthwise and layerwise pruning in terms of memory and FLOPs. Although the two methods remove parameters from different parts of the transformer structure, they yield nearly identical theoretical hardware costs when the overall compression ratio is held constant. For example, at 30\% compression on LLaVA, the difference in FLOPs between widthwise (6.89 T) and layerwise (6.92 T) pruning is less than 0.5\%.

However, measuring actual inference latency reveals a notable divergence. As shown in Table \ref{tab:model_size_combined}, layerwise pruning consistently yields lower inference latency than widthwise pruning across all tested models and compression ratios. For example, on LLaVA at 30\% compression, layerwise pruning achieves a latency of 80.7 ms, compared to 84.4 ms for widthwise pruning. This empirical advantage stems from fundamental differences in how each pruning strategy interacts with GPU execution. Layerwise pruning removes entire transformer blocks, directly reducing the number of sequential operations and eliminating multiple CUDA kernel launches per removed layer. Widthwise pruning also delivers meaningful speedups by reducing the size of matrix operations within each layer, but it preserves the original network depth, meaning the fixed overhead of launching kernels for every layer remains unchanged. Additionally, while smaller matrix dimensions reduce total computation, the relationship between matrix size and execution time is not strictly linear on modern GPUs due to memory bandwidth constraints and parallelization thresholds. 

Consequently, while both strategies provide immense memory and speed gains, layerwise pruning is more efficient for actual hardware execution. Nevertheless, because widthwise pruning better preserves reasoning capabilities (as discussed in Section \ref{Zero-Shot Performance.}), selecting the superior overall strategy requires a careful trade-off, balancing deployment needs for absolute latency reduction against the retention of task accuracy.

\begin{table*}[t]
\centering
\small
\setlength{\tabcolsep}{2pt}
\caption{Memory ($\text{GiB}$), FLOPS ($\text{T}$), and Latency($\text{ms}$) for Bunny, InternVL, and LLaVA at various compression ratios for widthwise and layerwise pruning}
\begin{tabular}{@{}c|ccc|ccc|ccc@{}}
\toprule
\multirow{2}{*}{Ratio} & \multicolumn{3}{c}{Bunny} & \multicolumn{3}{c}{InternVL} & \multicolumn{3}{c}{LLaVA} \\
\cmidrule{2-4} \cmidrule{5-7} \cmidrule{8-10}
 & Mem. & FLOPS & Latency & Mem. & FLOPS & Latency & Mem.  & FLOPS & Latency \\
\midrule
$0\%$   & $6.02$ & $4.77$ & $50 \pm 1.2$ & $8.22$ & $5.12$ & $58 \pm 1.5$ & $13.23$ & $9.57$ & $105 \pm 1.5$ \\
\midrule
\multicolumn{10}{c}{\textit{Widthwise Pruning}} \\
\midrule
$15\%$  & $5.25$ & $4.14$ & $47.2 \pm 1.1$ & $7.05$ & $4.41$ & $54.8 \pm 1.4$ & $11.26$ & $8.21$ & $98.2 \pm 3.1$ \\
$30\%$  & $4.49$ & $3.50$ & $43 \pm 1.0$ & $5.82$ & $3.73$ & $49.2 \pm 1.2$ & $9.32$ & $6.89$ & $84.4 \pm 2.5$ \\
$45\%$  & $3.68$ & $2.84$ & $38 \pm 0.8$ & $4.61$ & $3.06$ & $46 \pm 1.1$ & $7.29$ & $5.49$ & $64.5 \pm 1.9$ \\
$60\%$  & $2.92$ & $2.20$ & $32.3 \pm 0.6$ & $3.43$ & $2.38$ & $36.7 \pm 0.9$ & $5.31$ & $4.17$ & $55.4 \pm 1.4$ \\
\midrule
\multicolumn{10}{c}{\textit{Layerwise Pruning}} \\
\midrule
$15\%$  & $5.28$ & $4.16$ & $46.5 \pm 0.8$ & $7.16$ & $4.45$ & $54.3 \pm 1.1$ & $11.33$ & $8.03$ & $95 \pm 8.1$ \\
$30\%$  & $4.55$ & $3.56$ & $41.2 \pm 0.9$ & $6.11$ & $3.79$ & $48.7 \pm 1.3$ & $9.44$ & $6.92$ & $80.7 \pm 0.6$ \\
$45\%$  & $3.82$ & $2.95$ & $36.1 \pm 0.7$ & $5.05$ & $3.12$ & $42.3 \pm 0.9$ & $7.54$ & $5.55$ & $58.5 \pm 1.6$ \\
$60\%$  & $2.94$ & $2.22$ & $30.2 \pm 0.5$ & $3.55$ & $2.41$ & $32.4 \pm 0.8$ & $5.37$ & $3.90$ & $49.2 \pm 1.2$ \\
\bottomrule
\end{tabular}
\label{tab:model_size_combined}
\end{table*}

\subsection{Supervised Finetuning for Performance Recovery.}
\label{Supervised Finetuning for Performance Recovery After Pruning}
\begin{figure*}[t]
    \centering
    \includegraphics[width=\textwidth]{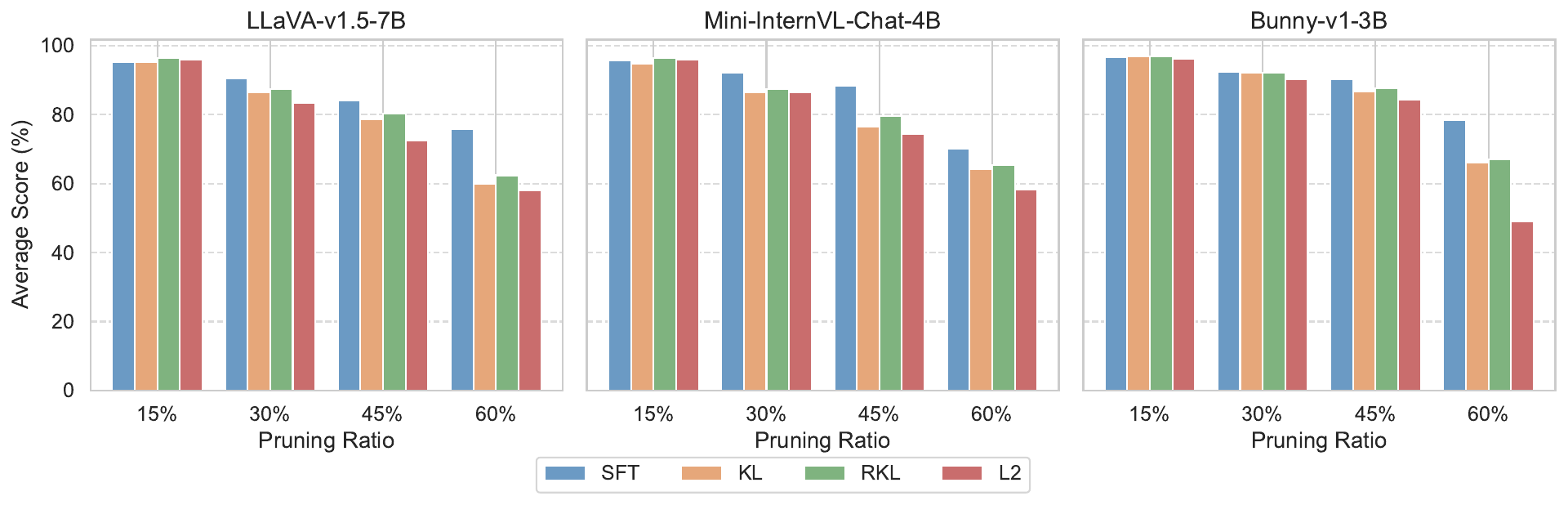}
    \caption{Performance comparison of KD strategies versus SFT across varying pruning ratios for LLaVA, InternVL, and Bunny models. KD achieves comparable results to SFT at low pruning ratios but performs significantly worse at higher ratios}
    \label{fig:kd_vs_sft}
\end{figure*}

\begin{figure*}[t]
    \centering
    \includegraphics[width=\textwidth]{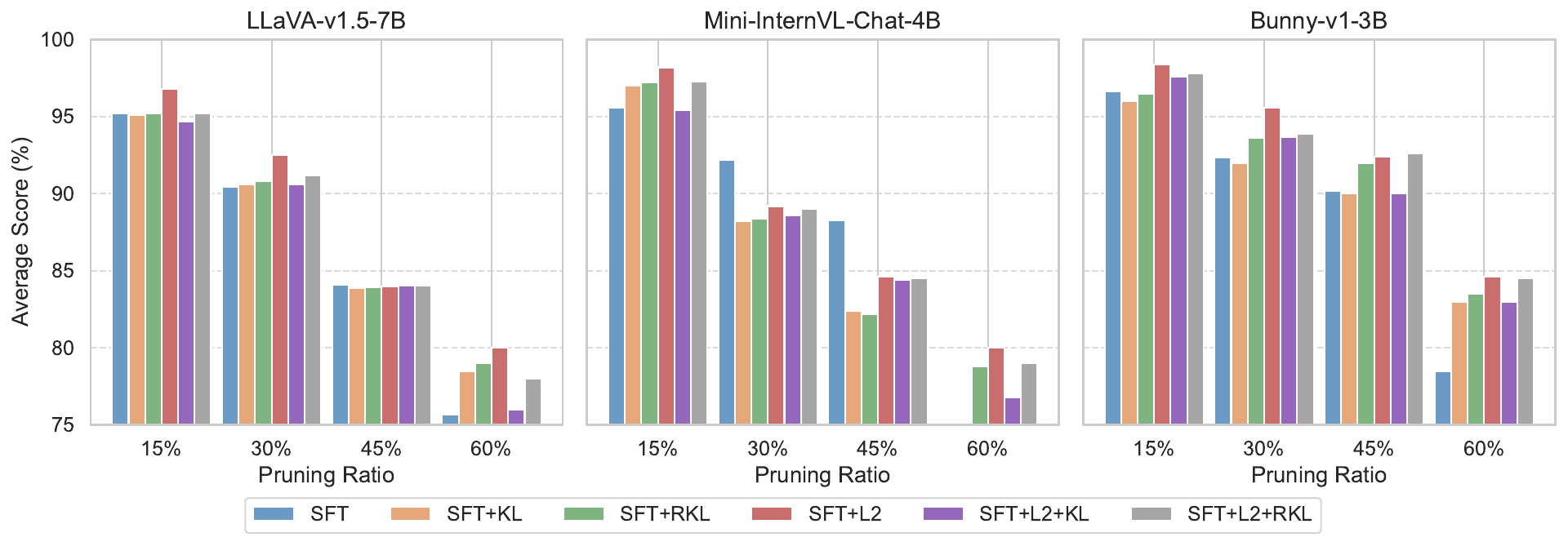}
    \caption{Performance comparison of combined recovery training techniques for LLaVA, InternVL, and Bunny models. Among the evaluated strategies, \textbf{SFT+L2} consistently demonstrates superior robustness and outperforms other loss combinations across varying pruning ratios}
    \label{fig:kd_plus_sft}
\end{figure*}
\begin{table}[t]
\centering
% \small
% \setlength{\tabcolsep}{6pt}
\caption{Disentangling multimodal versus language degradation. We compare average performance ($AVG$) and relative degradation ($Deg$) for LLaVA-v1.5-7B under widthwise pruning. Comparisons between multimodal ($mm$) and text-only ($txt$) inputs reveal that low compression ratios disproportionately affect modality alignment ($Deg_{mm}$) compared to language capabilities ($Deg_{txt}$)}
\begin{tabular}{ccccc}
\toprule
\textbf{Ratio} & \textbf{$AVG_{mm}$} & \textbf{$AVG_{txt}$} & \textbf{$Deg_{mm}$} & \textbf{$Deg_{txt}$} \\
\midrule
0.00\% & 53.70 & 40.30 & 0.00\% & 0.00\% \\
15.00\%  & 48.80 & 38.87 & 9.13\% & 1.07\% \\
30.00\% & 42.90 & 36.12 & 20.11\% & 10.38\% \\
45.00\% & 17.42 & 17.36 & 67.56\% & 56.92\% \\
60.00 \% & 3.62 & 2.80 & 93.26\% & 93.05\% \\
\bottomrule
\end{tabular}
\label{tab:mm_vs_txt}
\end{table}
Pruning the language backbones in LVLMs inevitably degrades their capabilities. To understand the mechanism of this degradation, we first disentangle the effects of pruning on modality alignment versus core language reasoning. We evaluate pruned models on the same benchmarks using both multimodal inputs and text-only inputs.  
Table~\ref{tab:mm_vs_txt} reveals a distinct disparity in performance degradation. At a low compression ratio of 15\%, multimodal performance drops significantly ($9.13\%$) while text-only performance remains largely intact ($1.07\%$ degradation). This implies that mild pruning primarily disrupts the \textit{alignment} between the vision and language components rather than the language model itself. However, as the compression ratio increases, both multimodal and text-only performance degrade sharply. This indicates that at higher pruning ratios, the degradation stems from damage to both the modality alignment and the fundamental language backbone.

To address these issues, we investigate two recovery strategies: (1) supervised finetuning~(SFT) only the multimodal projector, and (2) jointly SFT both the projector and the LLM. These experiments allow us to isolate the source of degradation and determine the most efficient recovery method. Following prior work suggesting that training the vision encoder can be detrimental~\cite{karamcheti2024prismatic}, we keep the vision encoder frozen in all setups. For the LLM, we employ Low-Rank Adaptation (LoRA)~\cite{hulora} to facilitate efficient finetuning.

\xhdr{SFT the multimodal projector.} 
As shown in Fig.~\ref{fig:finetune_plot} (blue lines), SFT the projector significantly restores performance, particularly at lower compression ratios. When the compression ratio is under $30\%$, updating the projector alone yields results comparable to jointly SFT the LLM. Notably, at a 15\% compression ratio, this approach retains over 95\% of the original performance for all the models. Even at a severe compression ratio of 60\%, projector SFT recovers 60--80\% of the original performance by re-aligning the visual features with the pruned language model. These results confirm that pruning induces a misalignment between visual and textual representations, which can be largely corrected by updating the projection layer.

\xhdr{SFT both the projector and the LLM.}
While projector SFT is effective for realignment, it cannot fully compensate for the loss of linguistic knowledge at higher compression rates. We observe consistent performance gains from additionally SFT the pruned LLM (red lines in Fig.~\ref{fig:finetune_plot}), specifically when the compression ratio exceeds 40\%. This aligns with our observation in Table~\ref{tab:mm_vs_txt} that the language backbone itself degrades significantly at these ratios. By SFT the LLM, we significantly restore these lost reasoning capabilities. For instance, at 45\% compression, joint SFT restores over 80\% of the original model's performance across all three models, underscoring its necessity when the language backbone is heavily compressed.

\textbf{\textit{\textcolor{blue!50!black!80}{Takeaway.}}}
When a small compression ratio of around 15\% is required, SFT the multimodal projector alone is typically sufficient to recover most of the model’s performance. For higher compression ratios ($\ge 45\%$), incorporating SFT of the LLM yields additional performance improvements.
\begin{figure*}[t]
    \centering
    \includegraphics[width=\linewidth]{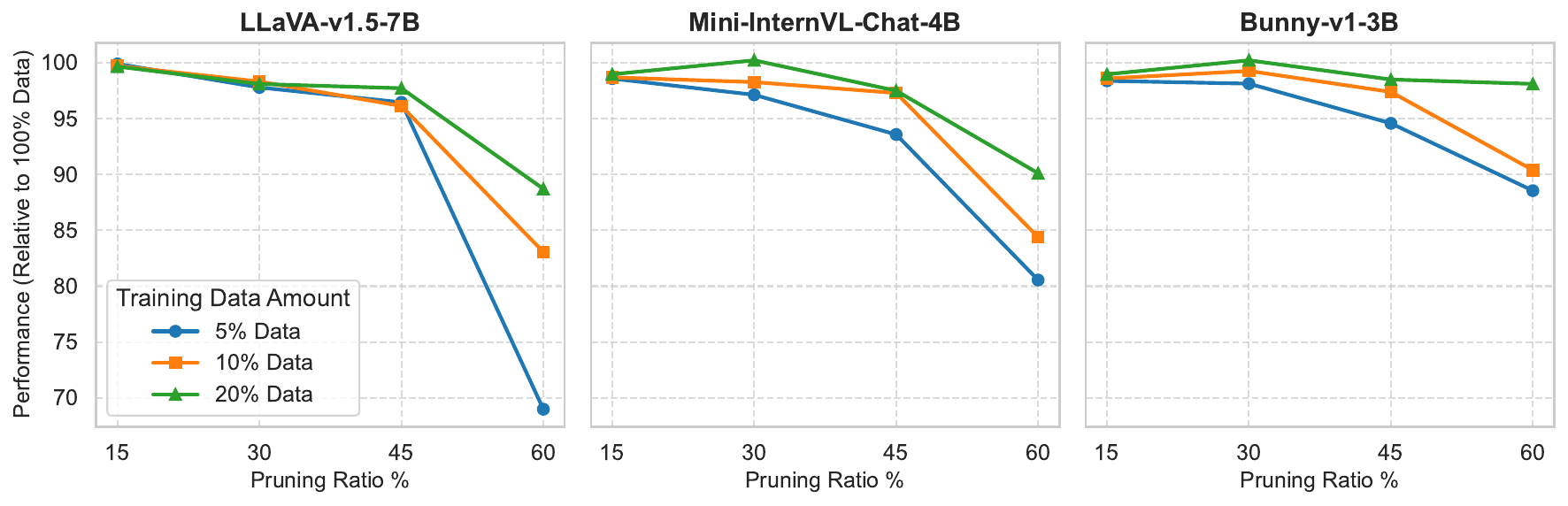}   
    \caption{Comparison of performance using different percentages of data for recovery training. For smaller compression ratios, even a small percentage of the training data (as low as 5\%) is sufficient to recover most of the original performance. However, as the compression ratio increases, more training data is required to achieve higher recovery performance
    }
    \label{fig:fewdata_plot}
\end{figure*}
\subsection{Knowledge Distillation for Performance Recovery.}
\label{kd_for_recovery}

Standard SFT recovers performance by retraining the model on ground-truth labels. However, the unpruned teacher model contains richer information—both in its output probability distributions and internal representations—that SFT ignores. We investigate whether distilling this knowledge back into the pruned student can further enhance recovery. We first compare standalone KD strategies against SFT, and then identify the optimal combination of SFT and KD.

\textbf{Comparison of KD strategies vs. SFT.} In Fig.~\ref{fig:kd_vs_sft}, we evaluate three KD objectives without SFT: standard KL Divergence (KL), Reverse KL (RKL), and L2 distance on hidden states (L2). 
At a low pruning ratio (15\%), all methods perform comparably to SFT. However, as the ratio increases, all KD methods fail to surpass SFT. Notably, L2 performance drops the fastest, becoming the least effective standalone strategy at high compression rates.

The degradation of KD-only methods underscores that without ground-truth labels (SFT), the pruned model lacks the strong supervision needed to realign its output space, especially at high pruning ratios. 
Regarding the specific KD trends ($RKL > KL > L2$), we attribute RKL's superiority over KL to its optimization behaviors. Standard KL is ``mean-seeking'': it forces the student to match the teacher's entire probability distribution, including low-probability tails. A heavily pruned model lacks the capacity to capture these nuances, resulting in uncertain predictions. 
In contrast, RKL is ``mode-seeking'': it penalizes the student only when it predicts high probability for an incorrect token, effectively encouraging the model to focus solely on the teacher's most likely prediction, as also noted in~\cite{gu2023knowledge}. For a capacity-constrained model, focusing on the single-best answer is more efficient than trying to mimic the teacher's full complexity.
Finally, L2 performs worst on its own because it aligns internal features but does not directly supervise the final classification head. Without output-level guidance, the model cannot effectively map these aligned features to correct text tokens.

\textbf{Synergy between SFT and KD.} 
To combine the benefits of hard labels and soft knowledge, we experiment with joint objectives in Fig.~\ref{fig:kd_plus_sft}, including dual combinations (e.g., SFT+KL, SFT+L2) and triple combinations (e.g., SFT+L2+KL).
Interestingly, the performance trend reverses: while L2 was the weakest standalone method, \textbf{SFT+L2} consistently outperforms all other strategies, including the more complex triple combinations.
This success stems from the orthogonality of the objectives: SFT optimizes the final output (the ``what'') while L2 acts as a regularizer for internal representations (the ``how''). Adding further output-level constraints (KL or RKL) on top of SFT+L2 yields diminishing returns or slight degradation, likely due to optimization conflicts between the hard labels and soft targets. Thus, the simple combination of SFT and feature alignment (L2) provides the most robust recovery.

\begin{table*}[t]
\begin{center}
\caption{Performance of the compressed models across various benchmarks. The Ratio indicates the proportion of LLM parameters removed. Layerwise pruning is applied for ratios below 45\%, while widthwise pruning is used for ratios 45\% and above.}
\resizebox{\linewidth}{!}{
\begin{tabular}{lrrrrrrrrrrrr}
\toprule

\multicolumn{1}{c}{\bf Method}
&\multicolumn{1}{c}{\bf Ratio}
&\multicolumn{1}{c}{\bf MMMU}
&\multicolumn{1}{c}{\bf SQA}
&\multicolumn{1}{c}{\bf MathVista}
&\multicolumn{1}{c}{\bf AI2D}
&\multicolumn{1}{c}{\bf MME-C}
&\multicolumn{1}{c}{\bf MME-P}
&\multicolumn{1}{c}{\bf POPE}
&\multicolumn{1}{c}{\bf DocVQA}
&\multicolumn{1}{c}{\bf GQA}
&\multicolumn{1}{c}{\bf AVG}
&\multicolumn{1}{c}{\bf AVG-\%}

\\ \midrule
\multicolumn{13}{c}{\bf LLaVA-v1.5-7B}
\\ \midrule[0.00005\arrayrulewidth]
- & 0.00\% & 35.10 & 68.67 & 26.70 & 54.80 & 363.21 & 1511.33 & 86.99 & 28.10 & 61.98 & 53.70 & 100.00\% \\
Layerwise+FT+L2 & 15.00\% & 36.40 & 68.42 & 25.50 & 53.70 & 337.86 & 1442.35 & 86.94 & 27.60 & 61.20 & 52.68 & 98.10\% \\
Layerwise+FT+L2 & 30.00\% & 36.00 & 68.82 & 23.40 & 50.50 & 318.57 & 1496.60 & 85.98 & 26.10 & 60.34 & 51.75 & 96.38\% \\
Widthwise+FT+L2 & 45.00\% & 30.80 & 52.92 & 20.20 & 48.40 & 215.00 & 1191.17 & 85.74 & 23.70 & 57.74 & 45.10 & 83.99\% \\
Widthwise+FT+L2 & 60.00\% & 27.70 & 46.26 & 18.00 & 46.30 & 211.79 & 1085.97 & 84.06 & 22.20 & 52.32 & 41.96 & 78.13\% \\
\midrule
\multicolumn{13}{c}{\bf InternVL-Chat-4B}
\\ \midrule[0.00005\arrayrulewidth]
- & 0.00\% & 43.20 & 93.30 & 53.70 & 76.90 & 547.50 & 1596.71 & 88.00 & 87.70 & 62.50 & 72.62 & 100.00\% \\
Layerwise+FT+L2 & 15.00\% & 42.80 & 92.47 & 53.22 & 76.21 & 542.61 & 1582.42 & 87.21 & 86.92 & 62.01 & 71.98 & 99.12\% \\
Layerwise+FT+L2 & 30.00\% & 41.45 & 89.53 & 51.53 & 73.79 & 525.36 & 1532.14 & 84.44 & 84.15 & 60.04 & 69.69 & 95.96\% \\
Widthwise+FT+L2 & 45.00\% & 37.24 & 80.43 & 46.29 & 66.30 & 472.00 & 1376.53 & 75.87 & 75.61 & 53.94 & 62.61 & 86.22\% \\
Widthwise+FT+L2 & 60.00\% & 34.29 & 74.06 & 42.63 & 61.05 & 434.62 & 1267.52 & 69.86 & 69.62 & 49.67 & 57.65 & 79.39\% \\
\midrule
\multicolumn{13}{c}{\bf Bunny-v1-3B}
\\ \midrule[0.00005\arrayrulewidth]
- & 0.00\% & 34.10 & 70.72 & 25.80 & 60.49 & 289.31 & 1487.71 & 87.82 & 22.80 & 54.72 & 51.89 & 100.00\% \\
Layerwise+FT+L2 & 15.00\% & 33.00 & 70.00 & 24.20 & 59.80 & 304.29 & 1457.06 & 87.97 & 21.80 & 54.56 & 51.36 & 98.98\% \\
Layerwise+FT+L2 & 30.00\% & 32.35 & 68.12 & 22.60 & 57.20 & 252.51 & 1349.91 & 87.53 & 20.80 & 53.08 & 48.97 & 94.37\% \\
Widthwise+FT+L2 & 45.00\% & 29.15 & 63.06 & 21.00 & 55.50 & 244.64 & 1281.66 & 87.09 & 19.80 & 52.31 & 46.95 & 90.48\% \\
Widthwise+FT+L2 & 60.00\% & 28.10 & 53.20 & 19.40 & 53.90 & 216.07 & 1115.33 & 86.73 & 18.80 & 48.72 & 43.51 & 83.86\% \\
\bottomrule
\end{tabular}
}
\label{tab:best_compressed_model}
\end{center}
\end{table*}
\textbf{\textit{\textcolor{blue!50!black!80}{{Takeaway.}}}}
Knowledge distillation, particularly when combined with SFT and using L2 loss to map the intermediate states, delivers the most effective performance recovery after pruning across all compression ratios.
%\end{tcolorbox}

% \input{figures/fewerdata_figure}
\subsection{Data Efficient Recovery.}
\label{Experiments_few_data}

We investigate the trade-off between training cost (data usage) and model performance. Fig.~\ref{fig:fewdata_plot} illustrates the recovery performance of widthwise-pruned models finetuned with varying fractions of the original dataset, using the optimal SFT+L2 strategy established in Section \ref{kd_for_recovery}.

For mild-to-moderate compression ratios \\
($\le 45\%$), we observe a rapid performance saturation. Remarkably, using just 5\% of the training data allows the model to recover over 90\% of the performance achieved with the full dataset. This trend holds consistently across different architectures. However, as the compression ratio increases beyond this point (e.g., 60\%), the performance gap between using partial data and full data widens, indicating that heavily pruned models require more extensive training examples to recover.

These findings suggest that the nature of the recovery task shifts with pruning severity. 
At lower compression ratios, the model retains most of its knowledge and structural integrity. In this regime, the recovery phase primarily serves as a calibration step that realigns the remaining weights and can be performed with a small, representative subset of data. 
In contrast, at high compression ratios, the model suffers significant capacity loss. The recovery process becomes a re-learning task, in which the model must discover new internal pathways to compensate for the removal of neurons. This requires a larger and more diverse set of examples to generalize effectively.

\textbf{\textit{\textcolor{blue!50!black!80}{{Takeaway.}}}}
With a compression ratio below 50\%, using only 5\% of the dataset is sufficient to achieve performance comparable to training on the full dataset. However, for compression ratios greater than 50\%, full-data training is necessary to recover performance effectively.

\begin{table*}[t]
\caption{Qualitative analysis of compressed LLaVA-v1.5-7B models following our established insights. \textbf{Bold text} indicates correctly identified visual details or counts, while \textit{italics} denote incorrect descriptions, hallucinations, or miscounts.}
\small
\centering
\begin{tabular}{r l p{6.8cm} p{4.8cm}}
\toprule
% Images and Questions aligned with responses
\multicolumn{2}{c}{} & 
\begin{minipage}[c]{6.8cm}
    \begin{minipage}[c]{0.35\linewidth}
        \includegraphics[width=\linewidth]{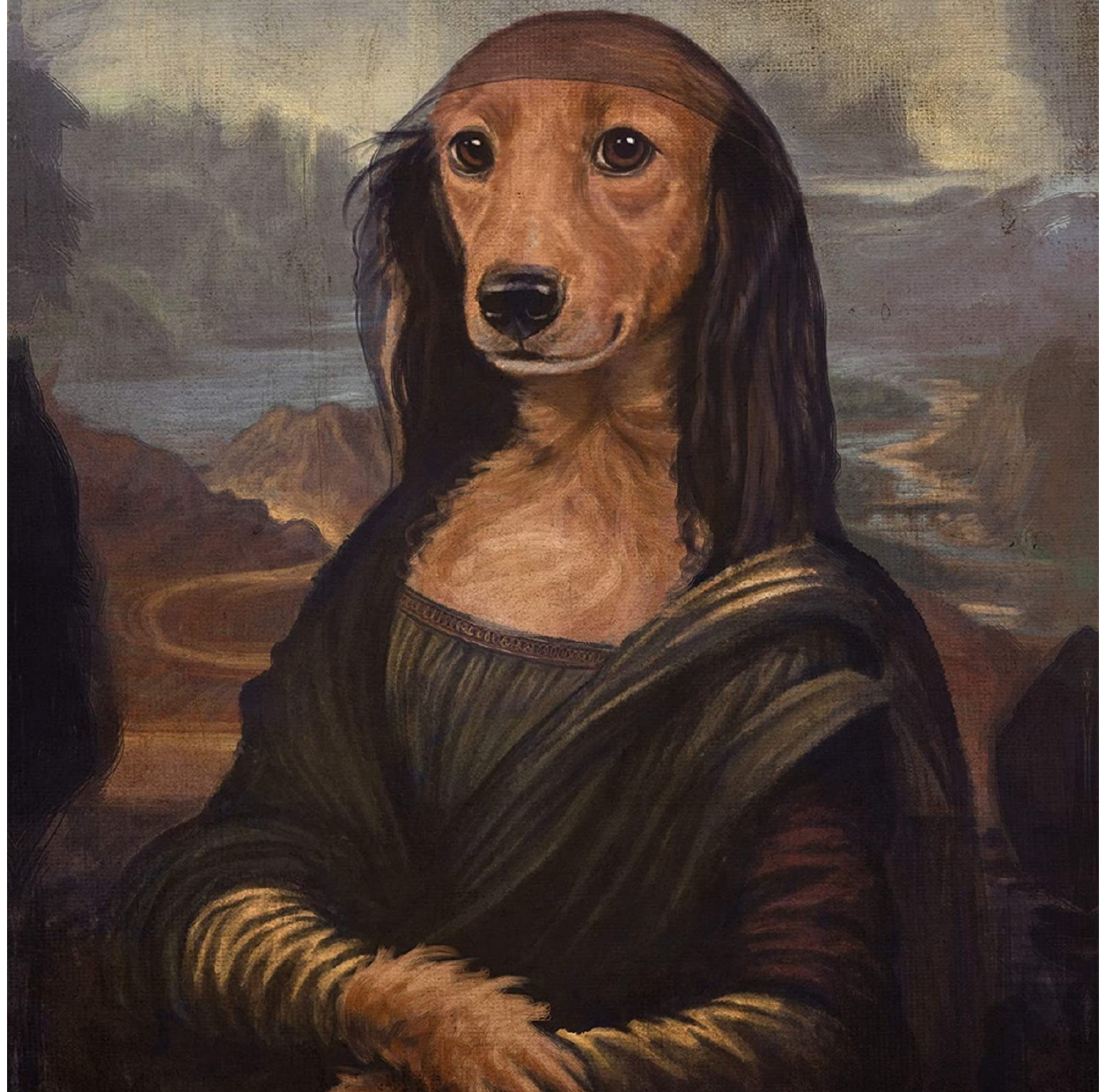}
    \end{minipage}%
    \hspace{0.03\linewidth}%
    \begin{minipage}[c]{0.45\linewidth}
        \textbf{Q1}: What is funny in the image?
    \end{minipage}
\end{minipage} & 
\begin{minipage}[c]{4.8cm}
    \begin{minipage}[c]{0.4\linewidth}
        \includegraphics[width=\linewidth]{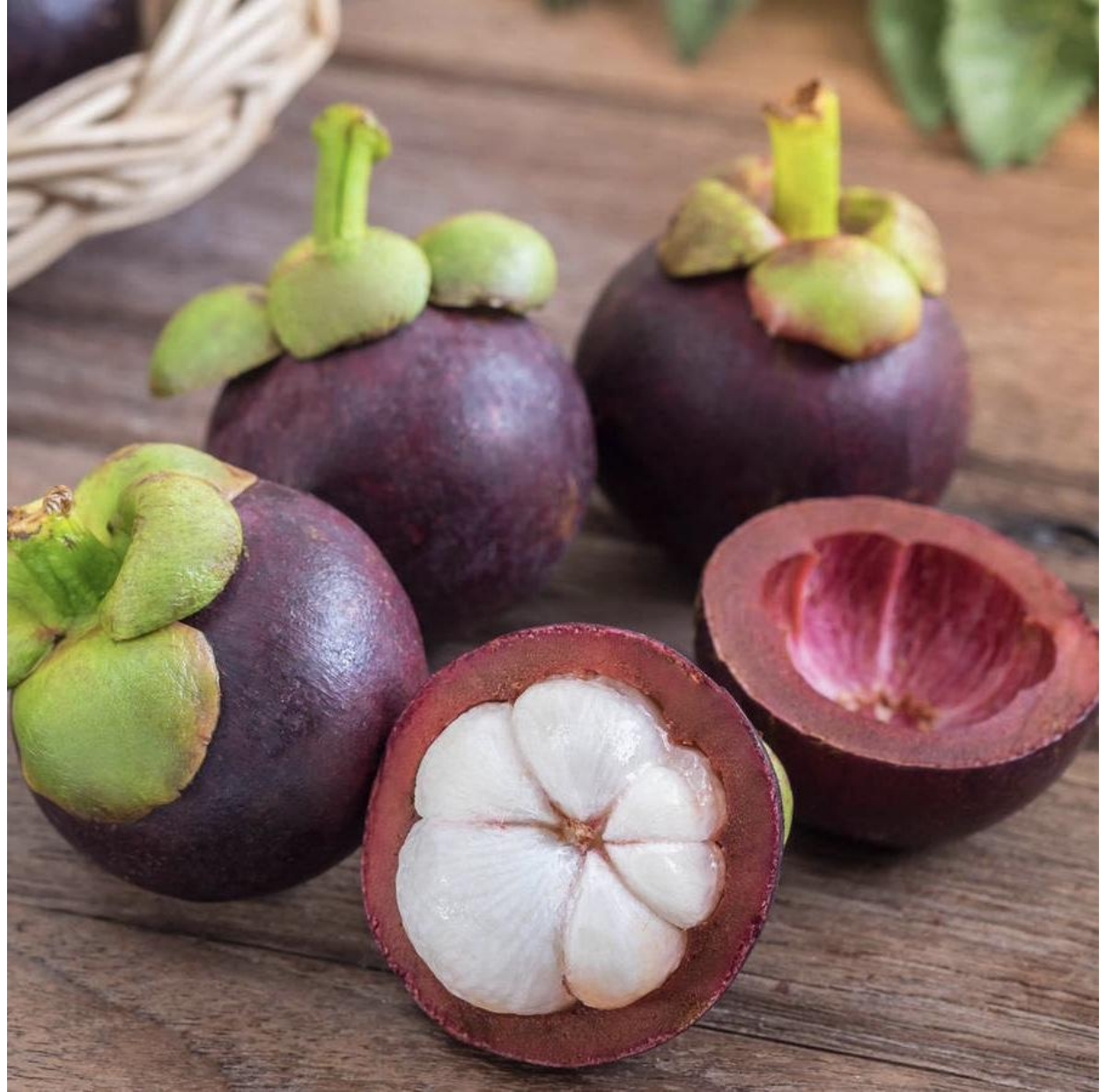}
    \end{minipage}%
    \hspace{0.04\linewidth}%
    \begin{minipage}[c]{0.45\linewidth}
        \textbf{Q2}: How many uncut fruits are in the image?
    \end{minipage}
\end{minipage} \\ 
\midrule
\textbf{Ratio} & \textbf{Strategy} & \textbf{Response 1} & \textbf{Response 2} \\ 
\midrule
$0\%$ & - &
The image is funny because it is \textbf{a parody of the famous Mona Lisa painting}, but with a \textbf{dog's head and paws} instead of a woman's. The dog is wearing the same \textbf{dark veil and dress} as seen in the original artwork. This is humorous because it replaces a serious and \textbf{iconic piece of art} with a cute dog, creating a silly and unexpected contrast. & 
There are \textbf{three uncut mangosteens} in the image. They are positioned in the background and to the left, showing their \textbf{dark purple skin and green stems}. Two other fruits have been cut open in the foreground to reveal the white flesh inside. \\
\midrule
$15\%$ & \makecell[l]{Layerwise \\ +FT+L2} &
The image is funny because it features \textbf{a dog wearing a Renaissance-style dress and a bandana}, posing as if it's \textbf{posing for a painting}. This is humorous because it is a playful representation of a dog, which is not something you would typically see in a \textbf{Renaissance-style dress}. & 
There are \textbf{three uncut fruits} in the picture. These are \textbf{mangosteens} with their \textbf{green caps} still attached. There are also \textit{three} other fruits that have been sliced open to show the inside. \\
\midrule
$30\%$ &\makecell[l]{Layerwise \\ +FT+L2}&
The image is funny because it features \textbf{a dog dressed up in a human-like attire}, wearing a \textbf{long black dress} and \textit{a hat}. This unusual scene is not something you would typically see in real life, as \textbf{dogs don't wear clothes} like humans. & 
There are \textbf{three uncut} \textit{purple onions} in the background. The image also shows some \textbf{fruit that has been cut in half} on a wooden surface. \\
\midrule
$45\%$ & \makecell[l]{widthwise \\ +FT+L2} & 
The image is funny because it features \textbf{a dog dressed in a human-like dress}, posing in a \textbf{pose as if it were a person}. The dog appears to be \textbf{dressed in a dress}, which is a humorous and unexpected scene, as dogs are not typically dressed like humans. \textbf{The dog's pose and the overall scene create a sense of humor and playfulness.} & 
There are \textit{two} uncut fruits. They look like \textit{pomegranates} or some kind of tropical fruit. \\
\midrule
$60\%$ & \makecell[l]{widthwise \\ +FT+L2}  & 
The image is funny because \textit{there is a woman sitting in a chair.} She is \textbf{wearing a black dress} and \textit{has long hair.} It is funny because \textit{she} has a \textbf{serious expression on \textit{her} face.} & 
There is \textit{one large purple cabbage} in the center. \textit{It is sitting on a wooden table next to some garlic.} The image shows a \textit{healthy vegetable harvest.} \\
\bottomrule
\end{tabular}
\label{tab:merged_qualitative}
\end{table*}

\subsection{Key Insights for Model Compression.}
\label{Model compression results following our best practices}
Based on the empirical results from the previous section, we outline the following practices for compressing LVLMs:

% \vspace{5pt}
\begin{tcolorbox}[colback=gray!10, colframe=blue!20!gray, rounded corners=all, before skip=0pt,
    left=0pt]
    \setlength{\leftmargini}{2pt}
    \begin{itemize}
        \item \textbf{Widthwise pruning is more effective}, yielding an efficient model even without recovery training.
        \item \textbf{With recovery training}, layerwise pruning excels for smaller compression ratios \
        ($\le 30\%$), while widthwise pruning performs better at higher ratios ($\ge 45\%$).
        \item \textbf{For small compression ratios} ($\le 15\%$), finetuning just the multimodal projector is often sufficient to restore performance.
        \item \textbf{For recovery training}, combining SFT with KD of the intermediate representations using L2 loss consistently achieves the highest performance.
        \item \textbf{Recovery is highly data-efficient,} requiring only 5\% of the original data to match full-data training results, though full datasets are still needed for high compression ratios. 
    \end{itemize}
\end{tcolorbox} 
   
Guided by these insights, we present the comprehensive performance benchmarks of our pruned and recovered models in Table \ref{tab:best_compressed_model}.

\textbf{Qualitative Analysis and Failure Cases.} Beyond numerical metrics, we validate the practical utility of our compressed and recovered models following our established insights through qualitative examples in Table \ref{tab:merged_qualitative}. For LLaVA-v1.5-7B models compressed by up to $45\%$ (retaining approximately 3.8B parameters), the results demonstrate that despite significant pruning, these models retain a strong capacity for understanding complex visual inputs. For instance, they effectively identify key elements in the first image, such as the dog’s attire, and generate contextually accurate descriptions of the humor. However, as the pruning ratio increases, fine-grained object recognition and counting gradually degrade. In the second image, while the model correctly counts three uncut fruits at $15\%$ and $30\%$ pruning ratios, it begins to exhibit semantic shifts, misidentifying the mangosteens as "purple onions" at $30\%$, and eventually miscounting them as "two pomegranates" at $45\%$. Furthermore, the analysis reveals a clear performance boundary at extreme pruning ratios (e.g., $60\%$). At this level, visual-semantic alignment degrades significantly: the model hallucinates a woman based on the global compositional structure (resembling the Mona Lisa) rather than local visual features, and completely fails the counting task by hallucinating a "large purple cabbage" and "garlic." This confirms that while our method effectively preserves broad capabilities across a wide range of sparsity levels, there is a critical threshold beyond which semantic granularity and precise visual grounding are compromised.

\section{Discussion}\label{Discussion}
In this section, we discuss the extent to which pruning the model makes sense (Sec. \ref{How much can we prune the model?}), and compare our best practices with quantization and explore their integration (Sec. \ref{Comparison and combination with quantization}). Finally, we discuss the limitations of structural pruning and recovery training as compression techniques, as well as potential directions for future work (Sec. \ref{Limitation and future work}).

\subsection{Impact of Calibration Dataset Size.}
\label{Impact of Calibration dataset size}
\begin{figure}[t]
    \centering
    \includegraphics[width=\linewidth]{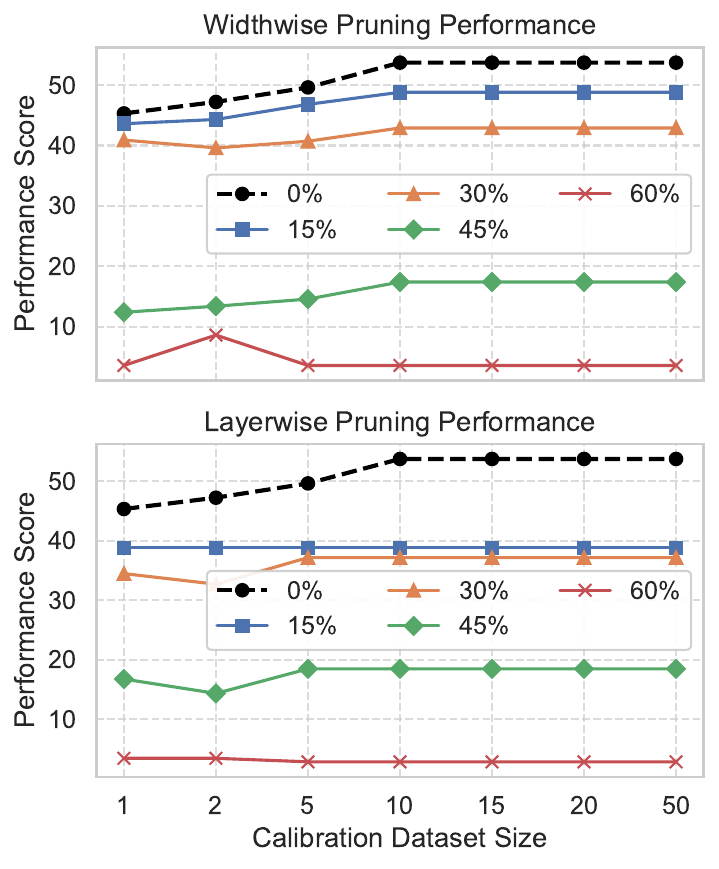}
    \caption{Impact of calibration dataset size on zero-shot performance (LLaVA-v1.5-7B). We evaluate model performance across varying pruning ratios (15\%–60\%) and paradigms (Widthwise vs. Layerwise) using calibration subsets ranging from 1 to 50 samples. The results demonstrate that pruning sensitivity is largely independent of calibration size, with importance scores converging rapidly
    }
    \label{fig:calibration_size}
\end{figure}

We investigate the sensitivity of the pruning process to the size of the calibration dataset. Fig.~\ref{fig:calibration_size} reports the performance of LLaVA-v1.5-7B under widthwise and layerwise pruning using calibration sets ranging from 1 to 50 samples. 

We observe a consistent trend where model performance improves as the calibration set size increases from 1 to 15 samples. However, beyond 10 samples, the performance effectively plateaus. For instance, with widthwise pruning at a 15\% ratio, the score rises from 43.60 (1 sample) to 48.80 (15 samples) and remains constant thereafter. This saturation point is consistent across different pruning ratios and strategies, suggesting that a very small subset of data is sufficient to calculate reliable importance scores.

The rapid stabilization of performance indicates that the ``importance'' of specific weights or layers is a structural property of the model rather than being highly dependent on specific data examples. In large pre-trained models, redundant neurons tend to remain inactive or contribute minimally across a wide variety of inputs. Once the calibration set spans a broad range of features, adding more samples does not significantly alter the ranking of parameters deemed ``safe'' to prune, allowing for highly data-efficient calibration.

\subsection{How much can we prune the model?}
\label{How much can we prune the model?}

A critical question in model compression is determining the maximum sparsity level that maintains acceptable performance. Table \ref{tab:best_compressed_model} summarizes the capabilities of LLaVA, InternVL, and Bunny across varying pruning ratios using our optimized recovery strategy.

\textbf{The Safe Zone ($\le 30\%$).}
Our results indicate that LVLMs exhibit significant redundancy. At a pruning ratio of 15\%, all models retain approximately 98--99\% of their original capabilities. Even at a 30\% compression ratio, the performance loss is minimal. This suggests that up to 30\% of the parameters in current LVLMs contribute little to the model's core reasoning and multimodal abilities.

\textbf{The Tipping Point (45\%).}
Performance degradation accelerates significantly once the pruning ratio exceeds 30\%. At 45\% compression, the retained performance for LLaVA and InternVL drops to 83.99\% and 86.22\%, respectively, falling below the 90\% fidelity threshold. While Bunny remains slightly more robust (retaining 90.48\%), the trend is clear: aggressive pruning beyond this point compromises the model's structural integrity.

 Consequently, if the goal is to preserve high fidelity ($>90\%$ performance), the compression ratio should generally be kept below 30\%. Within this range, pruning offers significant efficiency gains with negligible impact on multimodal reasoning; beyond it, users must accept a trade-off between higher efficiency and noticeable performance degradation.

\subsection{Combination with Quantization.}
\label{Comparison and combination with quantization}
While our study focuses primarily on structural pruning, integrating quantization into our framework can further optimize memory efficiency. Here, we demonstrate that pruning and quantization are highly complementary techniques. 

Using LLM.int8() \cite{dettmers2022gpt3} as a representative method, Table \ref{tab:quantization} shows that applying quantization to our already-pruned models yields compound memory savings. For instance, quantizing the 30\% pruned LLaVA-5B model reduces the total memory footprint to just 5.4 GiB (down from the 13.5 GiB baseline) while maintaining a strong average performance score of 50.83. This confirms that our structured pruning framework can be seamlessly paired with quantization to maximize memory efficiency without severe performance degradation.
\begin{table}[t]
\setlength{\tabcolsep}{9pt}
\caption{Memory usage and average performance of LLaVA-v1.5-7B when combining structural pruning (layerwise) and int8 quantization. The results demonstrate that these complementary techniques can be seamlessly combined to achieve compounded memory reductions.}
\centering
\begin{tabular}{lccc}
\toprule
Model & Memory & Ratio & Avg \\ 
 & (GiB) & & \\ 
\midrule
LLaVA-7B & 13.2 & 0\% & 53.7 \\ 
LLaVA-7B.int8() & 7.5 & 0\% & 52.5 \\ 
\midrule
LLaVA-6B & 11.3 & 15\% & 52.68 \\ 
LLaVA-6B.int8() & 6.5 & 15\% & 51.9 \\ 
\midrule
LLaVA-5B & 9.4 & 30\% & 51.75 \\ 
LLaVA-5B.int8() & 5.4 & 30\% & 50.8 \\ 
\bottomrule
\end{tabular}
\label{tab:quantization}
\end{table}

\subsection{Limitation and Future Work.}
\label{Limitation and future work}
Our experiments demonstrate the effectiveness of structural pruning with recovery training at moderate compression ratios (up to 30\%). 
However, our study has boundaries. First, we focused exclusively on structured pruning for hardware universality; we did not investigate unstructured or semi-structured methods~\cite{frantar2023sparsegpt,sunsimple}, which represent an orthogonal direction for compression requiring different hardware considerations. Second, we kept the vision encoder frozen. While motivated by the parameter imbalance shown in Table \ref{model architecture}, future holistic compression frameworks could explore pruning the vision backbone, particularly for architectures where the vision encoder is larger.
Finally, beyond this threshold, performance loss becomes increasingly difficult to recover, suggesting that for applications requiring more aggressive compression, the extreme pruning of a large model is not a viable approach.
Due to computational constraints, this work focuses on two pruning techniques applied to three different models. Future work could extend these findings to include a broader range of pruning techniques and models, further refining these strategies.  

\section{Conclusion}\label{conclusion}

We systematically evaluated two structural pruning schemes---widthwise and layerwise---on LLaVA-v1.5-7B, InternVL-Chat-4B, and Bunny-v1-3B, and paired them with lightweight recovery through supervised finetuning and knowledge distillation. 
From these experiments, we distilled a decision chart that guides practitioners in choosing the pruning route and recovery budget for different target compression ratios.
Our findings provide a concrete path to fit LVLMs within strict memory, compute, or energy budgets without surrendering performance.

\backmatter

\section*{Data Availability Statement}
The data used in this study are derived from publicly available datasets. The original visual instruction tuning datasets—LLaVA-v1.5-mix665k~\cite{liu2024improved}, InternVL-Chat-V1-2-SFT-Data~\cite{chen2024internvl}, and Bunny-695K~\cite{he2024efficient}, were utilized for pruning calibration and recovery training. Evaluation was conducted using the lmms-eval~\cite{lmms_eval2024} suite across several public benchmarks, including MMMU~\cite{yue2023mmmu}, SQA\cite{lu2022learn}, MathVista~\cite{lu2023mathvista}, AI2D, POPE\cite{li2023evaluating}, MME~\cite{yin2023survey}, GQA~\cite{hudson2019gqa}, and DocVQA~\cite{mathew2021docvqa}. The code and pruned model checkpoints will be made available on GitHub upon publication.

% \begin{appendices}

% An appendix contains supplementary information that is not an essential part of the text itself but which may be helpful in providing a more comprehensive understanding of the research problem or it is information that is too cumbersome to be included in the body of the paper.




% \end{appendices}


\bibliography{sn-bibliography}% common bib file
% \input sn-article.bbl

\end{document}